\documentclass[11pt]{article}

\pdfoutput=1

\usepackage{amsthm}
\usepackage{amsfonts}
\usepackage{amsmath}
\usepackage{amssymb}
\usepackage{arydshln}
\usepackage{accents}
\usepackage{algorithm}
\usepackage{algpseudocode}

\usepackage{blindtext}
\usepackage[english]{babel}

\usepackage{color}
\usepackage{cite}
\usepackage{caption}
\usepackage{cancel}

\usepackage{enumerate}
\usepackage{float}

\usepackage{geometry}
\usepackage{graphicx}

\usepackage{hyperref}

\usepackage{import}
\usepackage[utf8]{inputenc}

\usepackage{latexsym}
\usepackage{lscape}
\usepackage{mathtools}
\usepackage{multirow}
\usepackage{natbib}
  
\usepackage{pdflscape}
\usepackage{pdfpages}

\usepackage{relsize}
\usepackage{rotating}

\usepackage{soul}
\usepackage{subfiles}
\usepackage{subcaption}

\usepackage{times}
\usepackage{tikz}
\usepackage{tabstackengine}

\usepackage{xcolor}

\newcommand{\ut}[1]{\underaccent{\tilde}{#1}}
\renewcommand{\vec}[1]{\ut{#1}}

\def\0{\boldsymbol{0}}
\def\1{\boldsymbol{1}}
\def\2{\boldsymbol{2}}
\def\3{\boldsymbol{3}}
\def\4{\boldsymbol{4}}
\def\5{\boldsymbol{5}}
\def\6{\boldsymbol{6}}
\def\7{\boldsymbol{7}}
\def\8{\boldsymbol{8}}
\def\9{\boldsymbol{9}}

\def\Mu{\boldsymbol{\mu}}
\def\Alpha{\boldsymbol{\alpha}}
\def\DDelta{\boldsymbol{\Delta}}  
\def\Nu{\boldsymbol{\nu}} 

\def\trans{{\mathsmaller T}}

\theoremstyle{plain}
\newtheorem{thm}{Theorem}[section]
\newtheorem{theorem}[thm]{Theorem}

\newgeometry{left=1.2in, top=1.2in, right=1.2in, bottom=1.2in,nohead, includefoot, verbose, ignoremp}

\def\Fcal{\mathcal{F}}

\def\Hcal{\mathcal{H}}

\def\Xcal{\mathcal{X}}

\def\EE{\mathbb{E}}

\def\NN{\mathbb{N}}

\def\PP{\mathbb{P}}

\def\RR{\mathbb{R}}

\def\H{\mathbf{H}}

\def\S{\mathbf{S}}

\def\W{\mathbf{W}}
\def\X{\mathbf{X}}

\def\Z{\mathbf{Z}}

\def\i{\mathbf{i}}
\def\j{\mathbf{j}}

\def\u{\mathbf{u}}
\def\v{\mathbf{v}}
\def\w{\mathbf{w}}
\def\x{\mathbf{x}}

\bibpunct{(}{)}{,}{a}{}{,}
\title{The Geometry of Nonlinear Embeddings\\ in Kernel Discriminant Analysis}

\author{Jiae Kim\thanks{Jiae Kim is Graduate Student, Department
    of Statistics, The Ohio State University, 
    Columbus, OH 43210 (Email: kim.3887@osu.edu).}
   \and Yoonkyung Lee\thanks{
  Yoonkyung Lee is Professor, Department of Statistics, 
  The Ohio State University, 
  Columbus, OH 43210 (Email: yklee@stat.osu.edu).}
  \and Zhiyu Liang\thanks{Zhiyu Liang is Senior Staff Data Scientist, 
    Kohl's, 
    Milpitas, CA 95035 (Email: zhiyu.liang@kohls.com).}
}

\date{}

\begin{document}
\maketitle

\begin{abstract}
Fisher’s linear discriminant analysis is a classical method for
classification, yet it is limited to capturing linear features
only. Kernel discriminant analysis as an extension is known
to successfully alleviate the limitation through a nonlinear feature
mapping. We study the geometry of nonlinear embeddings in discriminant analysis with
polynomial kernels and Gaussian kernel by identifying the
population-level  discriminant function that depends on the data
distribution and the kernel. In order to obtain
the discriminant function, we solve a generalized
eigenvalue problem with between-class and within-class covariance 
operators. The polynomial discriminants are shown to capture the class
difference through the population moments explicitly.
For approximation of the Gaussian discriminant, we
use a particular representation of the Gaussian kernel by utilizing the
exponential generating function for Hermite polynomials. We also show
that the Gaussian discriminant can be approximated using randomized
projections of the data.
Our results illuminate how the data distribution and the kernel interact in
determination of the nonlinear embedding for discrimination, and
provide a guideline for choice of the kernel and its parameters.\\

\newcommand{\Keywords}[1]{\par\noindent
{\small{\em Keywords\/}: #1}}


\end{abstract}

\section{Introduction}

Kernel methods have been widely used in statistics and machine
learning for pattern recognition and analysis
\citep{Shawe-Taylor:Cristianini, learningkernel,kernel2008}.
They can be described in a unified framework with a special class of
functions called {\it kernels} encoding pairwise similarities between data
points. Such kernels enable nonlinear
extensions of linear methods seamlessly 
and allow us to deal with general types of data such as vectors, text
documents, graphs, and images.
Combined with problem-specific evaluation criteria typically in the form of a
loss function or a spectral norm of a kernel matrix, this kernel-based
framework can produce a variety of learning algorithms for regression,
classification, ranking, clustering, and dimension reduction. Popular
kernel methods include smoothing splines \citep{Gwahba}, support
vector machines \citep{vapnik}, kernel Fisher discriminant
analysis \citep{mika1999fisher, Baudatklda}, ranking SVM \citep{Joachims},
spectral clustering \citep{Scottfeature, tutorspeccluster}, and kernel
principal component analysis \citep{kpcascholkopf}. 

This paper regards the geometry of kernel discriminant analysis (KDA).
KDA is a nonlinear generalization of Fisher's linear discriminant analysis (LDA), which is
a standard multivariate technique for classification.
Intrinsically as a dimension reduction method, KDA looks for 
discriminants that embed multivariate data into a real line so that
decisions can be made easily in a low dimensional space.
For simplicity of exposition, we focus on the case of two classes.
Fisher's linear discriminant projects data along the direction that maximizes 
separation between classes. Extending this geometric idea,
kernel discriminant analysis finds a data embedding that maximizes the
ratio of the between-class variation to within-class variation
measured in the feature space specified by a kernel.
To determine the embedding as a discriminant, we solve a generalized
eigenvalue problem involving kernel-dependent covariance matrices.

We examine the kernel discriminant at the population level to illuminate
the interplay between the kernel
and the probability distribution for data. Of particular interest is
how the kernel discriminant captures the difference between two
classes geometrically, and how 
the choice of a kernel and associated kernel parameters affect the
discriminant in connection with salient features of the underlying distribution. 
As a continuous analogue of the kernel-dependent covariance matrices,
we define the between-class and within-class covariance
operators first and state the population version of
the eigenvalue problem using those operators
which depend on both the data distribution and the
kernel. For some kernels, we can obtain explicit
solutions and determine the corresponding population kernel discriminants.

Similar population-level analyses have been done for kernel PCA and spectral clustering 
  \citep{Zhugaussian, shi1, Liang:Lee2013}
  to gain insights into the interplay between the kernel and
  distributional features on low dimensional embeddings for  data visualization and
  clustering. 
  The population analyses of kernel PCA, spectral clustering, and KDA
  require a spectral analysis of kernel operators of different forms 
  depending on the method. They help us examine the dependence of eigenfunctions and
eigenvalues of the kernel operators on the data distribution,
which can guide applications of those methods in practice.

The population discriminants with polynomial kernels admit a
closed-form expression due to their finite dimensional feature map. 
Analogous to the geometric interpretation of Fisher's linear
discriminant that it projects data along the mean difference direction
after whitening the within-class 
covariance, the polynomial discriminants are characterized by the
difference in the population moments between classes.
By contrast, the Gaussian kernel does not allow a simple closed-form
expression for the discriminant because its feature map and associated
function space are infinite-dimensional.  We provide approximations
to the Gaussian discriminant instead using two representations of the
kernel. These approximations shed some light on the workings of KDA
with the Gaussian kernel.
By using a deterministic representation of the Gaussian kernel with the Hermite
polynomial generating function, we approximate the population
Gaussian discriminant with polynomial discriminants of degree as high
as desired for the accuracy of approximation. This implies that the
Gaussian discriminant captures the difference between classes through 
the {\it entirety} of the moments. Alternatively, using a stochastic representation of the
Gaussian kernel through Fourier features of random projections
\citep{rahimi2008random}, we can also view the Gaussian discriminant as an embedding
that combines the expected differences in sinusoidal features of
randomly projected data from two classes.

How are the forms of these population kernel discriminants related to the
task of minimizing classification error? 
To attain the least possible error rate,  
the optimal decision rule assigns a data point $\x\in \RR^p$ to the most probable
class by comparing the likelihood of one class, say $p_1(\x)$, versus the
other, $p_2(\x)$, given $\x$. In other words, the ideal data
embedding for discrimination of two classes should be based on the
likelihood ratio $p_1(\x)/p_2(\x)$  or  $\log [p_1(\x)/p_2(\x)]$.
As a simple example, when the population distribution for each class
is multivariate normal with a
common covariance matrix,  $\log [p_1(\x)/p_2(\x)]$ is linear in $\x$,
and it coincides with the population version of Fisher's linear discriminant.
Difference in the covariance brings additional quadratic terms to the
log likelihood ratio requiring a quadratic discriminant for the
lowest error. As the distributions further deviate from elliptical
scatter patterns exemplified by normal distributions, the
ideal data embedding according to $\log [p_1(\x)/p_2(\x)]$
will involve nonlinear terms beyond quadratic. The basic fact that each distribution can
be identified with its moment-generating function or characteristic
function, i.e., its Fourier transform, implies  
that any difference between two distributions can be described in terms of  
the moments or expected Fourier features in general.
Our population analysis of kernel discriminants indicates that 
the Gaussian kernel treats the distributional difference as
a whole, including both global and local (or low and high frequency)
characteristics, while the
polynomial kernels focus on differences in more global characteristics
represented by low-order moments. The ideal choice of a kernel
in KDA will inevitably depend on the mode of class
difference mathematically expressed through the log likelihood ratio, $\log [p_1(\x)/p_2(\x)]$.

The rest of the paper is organized as follows.
Section~\ref{sec:KDAintro} provides a brief review of kernel discriminant analysis
and describes its population version by introducing two kernel
covariance operators for measuring the between-class variation and
within-class variation in the feature space.
Section~\ref{sec:KDA} presents a population-level discriminant
analysis using two types of polynomial kernels and Gaussian kernel and provides
an explicit form of population kernel discriminants.  
Numerical examples are given in Section~\ref{sec:simstudy} to
illustrate the geometry of kernel discriminants in relation to the
data distribution. 
Section~\ref{sec:disc} concludes the paper with discussions.


\section{Preliminaries}\label{sec:KDAintro} 

This section provides a technical background for kernel discriminant analysis.
After reviewing kernel functions, corresponding function spaces, and feature mappings in Section~\ref{subsec:kernel}, we briefly describe Fisher's linear discriminant analysis and its extension using kernels in Section~\ref{subsec:kda} and further extend the sample-dependent description of kernel discriminant analysis to its population version in Section~\ref{subsec:popkda}.

\subsection{Kernel}\label{subsec:kernel}

Let the input domain for data be denoted by $\Xcal$.
A kernel $K(\cdot, \cdot)$ is defined as
a positive semi-definite function from $\Xcal \times \Xcal$ to $\RR$. 
As a positive semi-definite function,  $K$ is symmetric: $K(\x,\u) = K(\u,\x)$  for all $\x,\u \in \Xcal$, and for each $n\in \NN$ and  for all choices of $\x_1, \ldots, \x_n \in \Xcal$,  $K_n = [K(\x_i,\x_j)]$ as an $n\times n$ matrix is positive semi-definite.

Given $K$, there is a unique function space $\Hcal_K$ with inner product
$\langle \cdot, \cdot \rangle_{\Hcal_K}$ corresponding to the kernel such that
for every $\x \in \Xcal$ and $f \in \Hcal_K$, (i) $K(\x,\cdot) \in \Hcal_K$, and (ii) $f(\x) = \langle f, K(\x, \cdot)\rangle_{\Hcal_K}$. The second property is called the reproducing property of $K$, and it entails the following identity:
$K(\x,\u) = \langle K(\x,\cdot), K(\u, \cdot)\rangle_{\Hcal_K}$. 
Such a function space with reproducing kernel  is called a reproducing kernel Hilbert space (RKHS). See \cite{aronszajn1950, Gwahba} and \cite{gu2002smoothing} for reference.

Alternatively, kernels can be characterized as those functions that
arise as a result of  the dot product of feature vectors. This is a
common viewpoint  in machine learning in the use of kernels for nonlinear generalization of linear methods.
To capture nonlinear features often desired for data analysis, consider a mapping
$\boldsymbol{\phi}$ from the input space $\Xcal$ to a feature space $\Fcal=\RR^D$,   $\boldsymbol{\phi} : \Xcal \rightarrow \Fcal$, which is called a feature map. The feature vector $\boldsymbol{\phi}(\x)=(\phi_1(\x),\ldots,\phi_D(\x))^\trans$ consists of $D$ features,  and for expressiveness of the features, the dimension of the feature space is often much higher than the input dimension, and possibly infinite.
Through the dot product of feature vectors,
we can define a valid kernel $K$ on $\Xcal \times \Xcal$ as 
$K(\x,\u) =\boldsymbol{\phi}(\x)^\trans \boldsymbol{\phi}(\u)$.
When $D=\infty$, the dot product is to be interpreted in the sense of $\ell_2$ inner product.
More general treatment of kernels with a general inner product for the feature space is feasible, but for brevity, we confine our description to the dot product only.
Using a feature map, we can generalize a linear method by applying it
in the feature space, which amounts to replacing the dot product for
the original features, $\x^\trans \u$, in the linear method with a kernel, $K(\x,\u) $. This substitution is called the ``kernel trick” in machine learning. For general description of kernel methods, an explicit form of a feature map is not needed nor the feature map
for a given kernel is unique. See \cite{learningkernel} for general properties of kernels.  

In this paper, we focus on the following kernels that are commonly used in practice
with $\Xcal=\RR^p$: 
\begin{itemize}
\item Homogeneous polynomial kernel of degree $d$:
$K_d(\x,\u)=(\x^\trans \u)^d$
\item Inhomogeneous polynomial kernel  of degree $d$:
$\tilde{K}_d(\x,\u)=(1+\x^\trans \u)^d$
\item Gaussian kernel  with bandwidth parameter $\omega$:
$K_\omega(\x,\u)=\exp\left(-\frac{ \| \x-\u \|^2}{2\omega^2}\right)$. 
\end{itemize}
Consideration of their explicit feature maps will be useful for the analyses presented in Section~\ref{sec:KDA}. For instance, the homogeneous polynomial kernel of degree $2$ on $\Xcal = \RR^2$, $K_2(\x,\u) = (x_1u_1 + x_2u_2)^2$, can be described with
a feature map
$\boldsymbol{\phi}(\x)= \left(x_1^2, ~\sqrt{2}x_1x_2, ~ x_2^2 \right) ^\trans \in \RR^3$.
The Gaussian kernel on $\RR$ with bandwidth parameter 1 admits $\Fcal = \RR^\infty$ with $\ell_2$ inner product as a feature space and a feature map of
\[
  \boldsymbol{\phi}(x) = e^{-\frac{x^2}{2}} 
  \left( 1, ~ x, ~ \frac{x^2}{\sqrt{2!}}, ~\frac{x^3}{\sqrt{3!}}, ~\ldots \right)^\trans. 
\]

\subsection{Kernel Discriminant Analysis}\label{subsec:kda}

Kernel discriminant analysis (KDA) is a nonlinear extension of Fisher's linear discriminant analysis using kernels.
For description of KDA, we start with a classification problem.
Suppose we have data from two classes labeled 1 and 2: 
$\{(\x_i,y_i)~|~ \x_i \in \Xcal \mbox{ and } y_i\in\{1,2\}
\mbox{ for }i=1, \ldots,n\}$. For simplicity, assume that the data
points are ordered so that the first $n_1$ observations are from class
1 and the rest ($n_2=n-n_1$) are from class 2.

\subsubsection{Fisher’s Linear Discriminant Analysis}
As a classical approach to classification, Fisher’s linear
discriminant analysis (LDA) looks for linear combinations of the
original variables called {\it linear discriminants} that can separate
observations from different classes effectively. It can be viewed as a
dimension reduction technique for classification.

When $\Xcal=\RR^p$, a linear discriminant is of the form,
$f(\x)=\v^\trans\x$, with a coefficient vector $\v \in \RR^p$. For the
discriminant $\v^\trans\x$ as a univariate measurement, we define the between-class variation as
\[ (\v^\trans\overline{\x}_1- \v^\trans\overline{\x}_2)^2
  =\v^\trans (\overline{\x}_1- \overline{\x}_2)(\overline{\x}_1-\overline{\x}_2)^\trans\v\]
and the within-class variation as its pooled sample variance:
\[ \frac{n_1}{n} \v^\trans S_1\v+\frac{n_2}{n}\v^\trans S_2\v
  = \v^\trans \left(\frac{n_1}{n} S_1+\frac{n_2}{n}S_2 \right)\v,\]
where $\overline{\x}_j$ and $S_j$ are the sample mean vector and sample covariance matrix of $\x$ for class $j$. Letting
$S_B = (\overline{\x}_1- \overline{\x}_2)(\overline{\x}_1-\overline{\x}_2)^\trans$
and
$S_W=\frac{n_1}{n} S_1+\frac{n_2}{n}S_2$ (the pooled covariance matrix),
we can express the two variations succinctly as quadratic forms of
$\v^\trans S_B \v$ and $\v^\trans S_W \v$, respectively. Note that
both forms are shift-invariant.

To find the best direction that gives the maximum separation between
two classes measured relative to the within-class variance in LDA,  we
maximize the ratio of the between-class variation to the within-class
variation with respect to $\v$:
\[\underset{\v \in \RR^p}{\text{maximize}}
    \frac{\v^\trans S_B \v}{\v^\trans S_W \v}.\]
This ratio is also known as the {\it Rayleigh quotient} and taken as a
measure of  the signal-to-noise ratio in classification along the direction $\v$. 
This maximization problem leads to the following generalized eigenvalue problem: 
\[ S_B \v=\lambda S_W \v\]
and the solution is given by the leading eigenvector.
More explicitly, 
$\hat{\v} = S_W^{-1} (\overline{\x}_1-\overline{\x}_2)$ defined only up to a normalization constant, and 
$\hat{\lambda}=(\overline{\x}_1-\overline{\x}_2)^\trans S_W^{-1}
(\overline{\x}_1-\overline{\x}_2)$ is the corresponding eigenvalue.
Since $S_B$ has rank 1, $\hat{\lambda}$ is the only positive eigenvalue.  
The resulting linear discriminant, $\hat{f}(\x)=\hat{\v}^\trans \x$, together with an appropriately chosen threshold $c$ yields
a classification boundary of the form
$\{\x\in \RR^p~|~ \hat{\v}^\trans \x=c\}$, which is linear in the input space.
When $S_W \approx I_p$, $\hat{\v} \approx
\overline{\x}_1-\overline{\x}_2$ (mean difference) provides the best direction for
projection. Re-expression of the linear discriminant as 
$\hat{f}(\x)=\hat{\v}^\trans \x=\big[S_W^{-\frac{1}{2}}
(\overline{\x}_1-\overline{\x}_2)\big]^\trans S_W^{-\frac{1}{2}}\x$
further  reveals that LDA projects data onto the mean
difference direction after {\it whitening} the variables via $S_W^{-\frac{1}{2}}$.
This interpretation also implies the invariance of $\hat{f}(\cdot)$ under 
variable scaling.
 
\subsubsection{Nonlinear Generalization}

Using the aforementioned kernel trick, 
\cite{mika1999fisher} proposed a nonlinear extension of linear
discriminant analysis,  which can be useful when the optimal classification boundary is not linear.
Conceptually, by mapping the data into a feature space using a kernel,
kernel discriminant analysis finds the best direction for discrimination and corresponding linear discriminant in the feature space, which then defines a nonlinear discriminant function in the input space.

Given kernel $K$, let $\boldsymbol{\phi}: \Xcal \rightarrow {\cal F}$ be a feature map. Then using the feature vector $\boldsymbol{\phi}(\x)$, we can define the sample means and between-class and within-class covariance matrices in the feature space  analogously. These matrices are 
denoted by $S_B^{\boldsymbol{\phi}}$ and $S_W^{\boldsymbol{\phi}}$.  KDA aims to find $\v$ in the feature space that maximizes 
\begin{equation}
  \frac{\v^\trans S_B^{\boldsymbol{\phi}} \v}{\v^\trans S_W^{\boldsymbol{\phi}} \v}. \label{eq:kldacrit}
\end{equation}
When $\v$ is in the span of all training feature vectors $\boldsymbol{\phi}(\x_i)$, it can be expressed as $\v=\sum_{i=1}^{n} \alpha_i \boldsymbol{\phi}(\x_i)$ for some $\Alpha=(\alpha_1,\ldots,\alpha_n)^\trans \in \RR^n$.
When we plug $\v$ of the form into the numerator and denominator of the ratio in 
\eqref{eq:kldacrit} and expand both  
in terms of $\alpha_i$ using the kernel identity $K(\x,\u) = \boldsymbol{\phi}(\x)^\trans \boldsymbol{\phi}(\u)$, we have
\begin{equation*}
  \v^\trans S_B^{\boldsymbol{\phi}} \v = \Alpha^\trans B_n \Alpha \quad \mbox{and} \quad  \v^\trans S_W^{\boldsymbol{\phi}} \v = \Alpha^\trans W_n \Alpha, 
\end{equation*}
where $B_n$ and $W_n$ are the $n \times n$ matrices defined through the kernel 
that reflect between-class variation and within-class variation, respectively.
To describe $B_n$ and $W_n$ precisely, start with the kernel matrix
$K_n=[K(\x_i,\x_j)]$. It
can be partitioned into $[K_1 ~ K_2]$ with $n\times n_1$ matrix of
$K_1$ and $n\times n_2$ matrix of $K_2$,  according to the class label $y_i$.
Using this partition of $K_n$, we can show that
$B_n=(\bar{K}_1-\bar{K}_2) (\bar{K}_1-\bar{K}_2)^\trans$ with
$\bar{K}_j=\frac{1}{n_j} K_j 1_{n_j}$  and
\[W_n=\frac{n_1}{n}K_1\left(\frac{1}{n_1}I_{n_1}-\frac{1}{n_1^2}J_{n_1}\right)K_1^\trans+
\frac{n_2}{n} K_2\left(\frac{1}{n_2}I_{n_2}-\frac{1}{n_2^2}J_{n_2}\right)K_2^\trans,\]
where $1_{n_j}$ is the $n_j$ vector of ones, and $J_{n_j}=1_{n_j}1_{n_j}^\trans$ (${n_j}\times {n_j}$ matrix of ones). 

In order to find the best discriminant direction $\v=\sum_{i=1}^{n} \alpha_i \boldsymbol{\phi}(\x_i)$,
we maximize
$\displaystyle \frac{\Alpha^\trans B_n\Alpha}{\Alpha^\trans W_n\Alpha}$ with respect to
$\Alpha \in \RR^n$ instead. The solution is again given by the leading eigenvector of the generalized eigenvalue problem:
\begin{equation}
  B_n\Alpha = \lambda W_n \Alpha. \label{BalphaWalpha}
\end{equation}
Further, the estimated direction
$\hat{\v}=\sum_{i=1}^{n} \hat{\alpha}_i \boldsymbol{\phi}(\x_i)$ results in
the discriminant function of the form:
\begin{equation}
  \hat{f}(\x)=\hat{\v}^\trans \boldsymbol{\phi}(\x)
  =\sum_{i=1}^n \hat{\alpha}_i\boldsymbol{\phi}(\x_i)^\trans \boldsymbol{\phi}(\x) = \sum_{i=1}^n \hat{\alpha}_i K(\x_i,\x).\label{empiricalkdaembedding}
\end{equation}
Obviously $\hat{f}(\cdot)$ is in the span of $K(\x_i,\cdot)$, $i=1,\ldots, n$, and 
belongs to the reproducing kernel Hilbert space $\Hcal_K$.
As with Fisher's linear discriminant, the kernel discriminant function is 
determined only up to a normalization constant. To specify a decision rule completely, we need to choose an appropriate threshold for the discriminant function.

\subsection{Population Version of Kernel Discriminant Analysis}
\label{subsec:popkda}

To understand the effects of the data distribution, geometrical
difference between two classes, in particular, and the kernel on the resulting discriminant function, we consider a  population version of KDA.
For proper description of the population version, we first assume that 
$\{\x_1,\ldots,\x_n\}$ in the dataset is a random sample of $\X$ from a mixture of two distributions $\PP_1 $ and  $\PP_2 $ with  population proportions of $\pi_1$ and  $\pi_2(=1-\pi_1)$ for two classes, or $\PP= \pi_1 \PP_1 + \pi_2 \PP_2$.

To illustrate how the sample version of KDA extends to the population
version under this assumption, we begin with the eigenvalue problem in \eqref{BalphaWalpha}. 
Suppose $\lambda_n$ and $\Alpha = (\alpha_1, \ldots, \alpha_n)^\trans$ are a pair of eigenvalue and eigenvector satisfying \eqref{BalphaWalpha}.
After scaling both sides of \eqref{BalphaWalpha} by the sample size $n$, 
we have 
\begin{equation}
  \frac{1}{n}\sum_{j=1}^n B_n(i,j)\alpha_j
  = \frac{\lambda_n}{n} \sum_{j=1}^n W_n(i,j) \alpha_j
  \qquad \mbox{for} \quad i=1,\ldots,n.
  \label{discriteBW}
\end{equation}
As a continuous population analogue of $B_n$ and $W_n$, we can define the following bivariate functions on $\Xcal \times \Xcal$:
\begin{eqnarray}
B_K(\x,\u) &=& \biggl\{
\EE_1[K(\x,\X)] - \EE_2[ K(\x,\X) ] \biggl\}
  \biggl\{ \EE_1[ K(\u,\X) ] -  \EE_2[ K(\u,\X) ] \biggl\}\label{between} \\
 W_K(\x,\u) &=&\pi_1\text{Cov}_1 [ K(\x,\X), K(\u,\X)  ] + 
  \pi_2\text{Cov}_2 [ K(\x,\X), K(\u,\X)  ],\label{within}
\end{eqnarray}
where $\EE_j$ and $\text{Cov}_j$ indicate that the expectation and
covariance are taken  with respect to $\PP_j$. The matrices $B_n$ and
$W_n$ can be viewed as a sample version of $B_K(\cdot,\cdot)$ and
$W_K(\cdot,\cdot)$ evaluated at all pairs of data points $\x_1,
\ldots, \x_n$.

Further treating $\Alpha = (\alpha_1, \ldots, \alpha_n)^\trans$ as a discrete version of a function $\alpha(\cdot)$ at the data points, i.e., $\Alpha=(\alpha(\x_1), \ldots, \alpha(\x_n))^\trans$, and taking the sample class proportion, $(n_j/n)$, as an estimate of the population proportion,  $\pi_j$, and $\lambda_n$ as a sample version of the population eigenvalue $\lambda$, we arrive at the following integral counterpart of \eqref{discriteBW}: 
\begin{equation}
  \int_{\mathcal{X}} B_K(\x,\u)\alpha(\u) \, d\PP(\u) = \lambda
  \int_{\mathcal{X}} W_K(\x,\u)\alpha(\u)\, d\PP(\u)
  \quad \mbox{ for every }\x \in \Xcal.
  \label{eq:kldafcn}
\end{equation}
This eigenvalue problem involves two integral operators:
(i) the between-class covariance operator defined as
\[ {\cal B}[\alpha(\x)]=\int_{\cal X} B_K(\x,\u) \alpha(\u) d\PP(\u),\]
and (ii) the within-class covariance operator defined as 
\[ {\cal W}[\alpha(\x)]=\int_{\cal X} W_K(\x,\u) \alpha(\u) d\PP(\u).\]
The form of the sample discriminant function in \eqref{empiricalkdaembedding}
with scaling of $1/n$ suggests that using the solution to equation
\eqref{eq:kldafcn}, $\alpha(\cdot)$, 
we define the population discriminant function as 
\begin{equation}
 f(\x) = \int_\mathcal{X} K(\x,\u)\alpha(\u)  \, d\PP(\u).\label{eq:kldadisc}
\end{equation} 

Clearly, the eigenfunction $\alpha(\cdot)$ depends on the kernel $K$ and probability distribution $\PP$, and so does the kernel discriminant function with
$\alpha(\cdot)$  as a coefficient function. Hence, identification of the solution to 
the generalized eigenvalue problem in \eqref{eq:kldafcn} 
will give us better understanding of kernel discriminants in relation
to the data distribution and the choice of the kernel.
The correspondence between the pattern of class difference and the nature of the resulting discriminant is of particular interest.

\section{Kernel Discriminant Analysis with Covariance Operators} 
\label{sec:KDA}

In this section, we carry out a population-level  discriminant analysis with two types of polynomial
kernels and Gaussian kernel and derive an explicit form of population discriminant functions. 
Section~\ref{subsec:generalKDAPoly} covers the case with polynomial
kernels in $\RR^p$. Section~\ref{subsec:GaussianKDA} extends it 
to the Gaussian kernel using two types of kernel representations.

\subsection{Polynomial Discriminant}
\label{subsec:generalKDAPoly}

Starting with $\Xcal=\RR^2$, we lay out steps
necessary for a population version of discriminant analysis
with homogeneous polynomial kernel and derive the
population kernel discriminant function in Section
\ref{Homopolyd2}. We then extend the results to a multi-dimensional
setting with homogeneous polynomial kernel in
Section~\ref{Homopolydp} and inhomogeneous polynomial in Section~\ref{InHomopolydp}.

\subsubsection{Homogeneous Polynomial Kernel in Two-Dimensional Setting}\label{Homopolyd2}

The homogeneous polynomial kernel of degree $d$ in $\RR^2$ is 
\begin{eqnarray}
  K_d(\x,\u)&=& \left(x_1u_1 + x_2 u_2\right)^d = \sum_{i=0}^d {d \choose i}(x_1u_1)^{d-i}(x_2u_2)^{i} = \sum_{i=0}^d {d \choose i}\left(x_1^{d-i} x_2^{i}\right) \left( u_1^{d-i}u_2^{i} \right).
\label{twodhomopoly}
\end{eqnarray}
The simple form of $K_d$ allows us to obtain
the between-class variation function $B_K(\x,\u)$ in \eqref{between} and within-class variation function $W_K(\x,\u)$ in \eqref{within} explicitly in terms of the population parameters.

For $B_K(\x,\u)$, we begin with
\begin{eqnarray*}
 && \EE_1[K_d(\x,\X)] - \EE_2[ K_d(\x,\X) ] \\ 
&=& \EE_1\left[\sum_{i=0}^d {d \choose i }\left(x_1^{d-i} x_2^{i}\right) \left(X_1^{d-i} X_2^{i}\right)\right]-\EE_2\left[\sum_{i=0}^d {d \choose i}\left(x_1^{d-i} x_2^{i}\right) \left(X_1^{d-i} X_2^{i}\right)\right]\\
&=& \sum_{i=0}^d {d \choose i } \left(x_1^{d-i} x_2^{i}\right) \left(
    \EE_1[ X_1^{d-i} X_2^{i}] -\EE_2[ X_1^{d-i} X_2^{i}]\right),
\end{eqnarray*}
which depends on the difference in the moments of total degree $d$ between two classes.
Letting $\Delta_i = \EE_1[ X_1^{d-i} X_2^{i}] -\EE_2[ X_1^{d-i} X_2^{i}]$, the difference in moments, we can express $B_K(\x,\u)$ as 
\begin{eqnarray}
B_K(\x,\u)&=& \biggl\{
\EE_1[K_d(\x,\X)] - \EE_2[ K_d(\x,\X) ] \biggl\}
  \biggl\{\EE_1[ K_d(\u,\X) ] -  \EE_2[ K_d(\u,\X) ] \biggl\} \nonumber\\
&=&\left\{ \sum_{i=0}^d {d \choose i } \left(x_1^{d-i} x_2^{i}\right) \Delta_i  \right\} \left\{ \sum_{j=0}^d {d \choose j } \left(u_1^{d-j} u_2^{j}\right) \Delta_j  \right\}\nonumber\\
&=& \sum_{i=0}^d \sum_{j=0}^d {d \choose i }{d \choose j } \Delta_i\Delta_j  \left(x_1^{d-i} x_2^{i}\right)\left(u_1^{d-j} u_2^{j}\right).\nonumber
\end{eqnarray}

Similarly, for $W_K(\x,\u)$, using the form of $K_d$, we first derive the covariance for each class ($l=1,2$)
\begin{eqnarray*}
\text{Cov}_l[K_d(\x,\X),K_d(\u,\X)] &= & \text{Cov}_l\left[
 \sum_{i=0}^d {d \choose i }\left(x_1^{d-i} x_2^{i}\right) \left(X_1^{d-i} X_2^{i}\right),\sum_{j=0}^d {d \choose j }\left(u_1^{d-j} u_2^{j}\right) \left(X_1^{d-j} X_2^{j}\right)\right]\\
&=& \sum_{i=0}^d\sum_{j=0}^d  {d \choose i } {d \choose j }\left(x_1^{d-i} x_2^{i}\right)\left(u_1^{d-j} u_2^{j}\right) \text{Cov}_l[X_1^{d-i} X_2^{i},X_1^{d-j} X_2^{j}].
\end{eqnarray*}
Letting $ W_{i,j} = \pi_1 \text{Cov}_1[X_1^{d-i} X_2^{i},X_1^{d-j} X_2^{j}]+\pi_2 \text{Cov}_2[X_1^{d-i} X_2^{i},X_1^{d-j} X_2^{j}]$, the within-class covariance of
a pair of polynomial features of degree $d$, 
we can express the within-class variation function as 
\begin{eqnarray*}
W_K(\x,\u) &=&\pi_1\text{Cov}_1 [ K_d(\x,\X),K_d(\u,\X)  ] + 
  \pi_2\text{Cov}_2 [ K_d(\x,\X),K_d(\u,\X)  ]\\
&=& \sum_{i=0}^d\sum_{j=0}^d  {d \choose i } {d \choose j } W_{i,j}\left(x_1^{d-i} x_2^{i}\right)\left(u_1^{d-j} u_2^{j}\right). 
\end{eqnarray*}

Using these two functions for $K_d$, we obtain
the between-class covariance operator as 
\begin{eqnarray*}
\int_{\RR^2} &&  \hspace{-1cm} B_K(\x,\u) \alpha(\u) \, d\PP(\u) \\
&=& \int_{\RR^2}\sum_{i=0}^d \sum_{j=0}^d {d \choose i }{d \choose j } \Delta_i \Delta_j \left(x_1^{d-i} x_2^{i}\right)\left(u_1^{d-j} u_2^{j}\right)\alpha(\u)\, d\PP(\u) \\
&=& \sum_{i=0}^d \sum_{j=0}^d {d \choose i }{d \choose j } \Delta_i \Delta_j \left(x_1^{d-i} x_2^{i}\right)\int_{\RR^2}\left(u_1^{d-j} u_2^{j}\right)\alpha(\u)\, d\PP(\u)
\end{eqnarray*}
and the within-class covariance operator as 
\begin{eqnarray*}
\int_{\RR^2} && \hspace{-1cm} W_K(\x,\u) \alpha(\u) \, d\PP(\u) \\
&=& \int_{\RR^2}\sum_{i=0}^d \sum_{j=0}^d {d \choose i }{d \choose j }  W_{i,j} \left(x_1^{d-i} x_2^{i}\right)\left(u_1^{d-j} u_2^{j}\right)\alpha(\u)\, d\PP(\u)\\
&=& \sum_{i=0}^d \sum_{j=0}^d {d \choose i }{d \choose j }  W_{i,j} \left(x_1^{d-i} x_2^{i}\right)\int_{\RR^2}\left(u_1^{d-j} u_2^{j}\right)\alpha(\u)\, d\PP(\u).
\end{eqnarray*}
Given $\alpha(\u)$, $\int_{\RR^2}\left(u_1^{d-j} u_2^{j}\right) \alpha(\u) \, d\PP(\u)$ is a constant.
Thus, letting $\nu_j = {d \choose j }
\int_{\RR^2}\left(u_1^{d-j} u_2^{j}\right) \alpha(\u) \, d\PP(\u) $,
we arrive at the following eigenvalue problem from \eqref{eq:kldafcn}  for identification of $\alpha(\cdot)$:
\begin{equation*}
  \sum_{i=0}^d \sum_{j=0}^d {d \choose i } \Delta_i \Delta_j \nu_j\left(x_1^{d-i} x_2^{i}\right) = \lambda\sum_{i=0}^d\sum_{j=0}^d  {d \choose i }  W_{i,j}\nu_j\left(x_1^{d-i} x_2^{i}\right), 
\end{equation*}
which should hold for all $\x =(x_1,x_2)^\trans \in \RR^2$.
Rearranging the terms in the polynomial equation, we have
\begin{equation*}
  \sum_{i=0}^d \left\{ {d \choose i }\Delta_i    \sum_{j=0}^d  \Delta_j  \nu_j \right\} \left(x_1^{d-i} x_2^{i}\right)
  = \lambda\sum_{i=0}^d \left\{  {d \choose i } \sum_{j=0}^d W_{i,j}\nu_j\right\}  \left(x_1^{d-i} x_2^{i}\right). 
\end{equation*}
Matching the coefficients of $x_1^{d-i}x_2^{i}$ on both sides of the equation
leads to the following system of linear equations for $\Nu = (\nu_0, \ldots, \nu_d)^\trans$:
\begin{equation}
\DDelta \DDelta^\trans\Nu = \lambda  W\Nu,\label{eq:polyKDAeigen}
\end{equation}
where $\DDelta = (\Delta_0, \Delta_1, \ldots, \Delta_d)^\trans$ is a vector of the mean differences of $X_1^{d-i}X_2^i$ for $i=0,\ldots,d$, and $W = [W_{i,j}]$ is a weighted average of their covariance matrices.

When $d=1$,  $K_d$ becomes a linear kernel, and the features are just $X_1$ and $X_2$.
Thus, $\DDelta=\Mu_1-\Mu_2$ (population mean difference) and $W=\pi_1\Sigma_1 +\pi_2\Sigma_2$ (pooled population covariance matrix). 
Clearly, the eigenvalue problem in \eqref{eq:polyKDAeigen} reduces to that for the population version of Fisher's LDA  when $d=1$.

Assuming that $W^{-1}$ exists, we can show that the largest eigenvalue satisfying
equation \eqref{eq:polyKDAeigen} is $\lambda^*=\DDelta^\trans
W^{-1}\DDelta$ with eigenvector of 
$\Nu^* =W^{-1} \DDelta$. Given the best direction
$\Nu^* = (\nu_0, \ldots, \nu_d)^\trans $, the
population discriminant function $f(\cdot)$ in \eqref{eq:kldadisc} with
homogenous polynomial kernel of degree $d$ is
\begin{eqnarray*}
  \hspace{-.5cm} f(\x)
  &=&  \int_{\RR^2}  K_d(\x,\u)\, \alpha(\u) \, d\PP(\u)
       =\int_{\RR^2} \sum_{j=0}^d {d \choose j} x_1^{d-j}x_2^{j} u_1^{d-j}u_2^{j}\,\alpha(\u) \, d\PP(\u)\\
  &=&\sum_{j=0}^d x_1^{d-j}x_2^{j} \underbrace{{d\choose j}
      \int_{\RR^2}  u_1^{d-j}u_2^{j} \,\alpha(\u) \, d\PP(\u)}_{\nu_j}
      =\sum_{j=0}^d  \nu_j  x_1^{d-j}x_2^{j}.
\end{eqnarray*} 
We see that this polynomial discriminant  is expressed as a linear combination of the corresponding polynomial features and their coefficients are determined through the mean differences and variances of the features.

\subsubsection{Homogeneous Polynomial Kernel in Multi-Dimensional Setting}\label{Homopolydp}

We extend the result in $\Xcal=\RR^2$ to general $\RR^p$.
The homogeneous polynomial kernel of degree $d$ in $\RR^p$ is given as
\[
  K_d(\x,\u)=(\x^\trans \u)^d = \left(\sum_{i=1}^p x_iu_i\right)^d
  = \sum_{j_1+\cdots+j_p=d} {d \choose j_1,\ldots, j_p} \prod_{k=1}^p(x_ku_k)^{j_k}. 
\]
As a function of $\x$, it involves polynomials  in $p$ variables  of total degree $d$. To facilitate similar derivations as in $\RR^2$,  we will use a multi-index for the polynomial features.

Let $\j_d$ denote a $p$-tuple multi-index with non-negative integer entries that sum up to $d$. That is, $\j_d \in
\S_d := \{(j_1,\ldots,j_p) ~|~ j_i \in \NN\cup\{0\},  \sum_{i=1}^p j_i = d\}$ with cardinality of ${d+p-1 \choose d}$.
We will omit the subscript $d$ from $\j_d$ for brevity whenever it is clear from the context. 
For $\j = (j_1,\ldots,j_p) \in \S_d$, we abbreviate  the multinomial coefficient $ {d \choose j_1,\ldots, j_p} $ to ${d \choose \j}$, and  let $|\j|= j_1 + \cdots + j_p$ and
$\j!= \prod_{k=1}^p j_k!$. 
For $\x \in \RR^p$ and $\j \in \S_d$, let $\x^{\j}=x_1^{j_1}\cdots
x_p^{j_p}$, and  for $a \in \RR$, $a^\j$ means $ a^{j_1}\cdots a^{j_p}
= a^{|\j|}$. For convenience, we will use $\j \in \S_d$ and $|\j|=d$ interchangeably. 

Using this multi-index, we rewrite the homogeneous polynomial kernel in $\RR^p$ simply as
\begin{equation}
K_d(\x,\u) = \sum_{|\j|=d} {d \choose \j}\x^\j \u^\j,    \label{eq:polyK_d0}
\end{equation}
which can be viewed as a multi-dimensional extension of the expression in \eqref{twodhomopoly}. Further, we can derive the between-class and within-class variation functions similarly:
\begin{eqnarray*}
B_K(\x,\u)&=&\sum_{|\i|=d}\sum_{|\j|=d} {d \choose \i }  {d \choose \j }\Delta_\i \Delta_\j \x^\i\u^\j  \\ 
W_K(\x,\u)&=& \sum_{|\i|=d}\sum_{|\j|=d} {d \choose \i } {d \choose \j } W_{\i,\j}\x^\i \u^\j  
\end{eqnarray*}
with $\Delta_\i = \EE_1[ \X^\i]- \EE_2[ \X^\i]$ and
$W_{\i,\j} = \pi_1 \text{Cov}_1[\X^\i ,\X^\j]+\pi_2 \text{Cov}_2[\X^\i ,\X^\j]$
for $\i, \j \in \S_d$. As an example, when the degree $d$ is 2 in $\RR^3$,
$ \S_2 = \{(2,0,0),(1,1,0), (1,0,1), (0,2,0), (0,1,1), (0,0,2) \} $.
For $\i = (1,1,0)$ and $\j=(0,1,1)$, $\X^\i=X_1X_2$ and $\X^\j=X_2X_3$, and
thus we have 
$\Delta_\i = \EE_1[X_1X_2] - \EE_2[X_1X_2]$ and
$W_{\i,\j} =  \pi_1\text{Cov}_1 [ X_1X_2, X_2X_3] + \pi_2\text{Cov}_2[ X_1X_2, X_2X_3]$.
Due to the same structure, we can easily extend the between-class
and within-class covariance operators.

To identify the population discriminant function in this setting, we define
$\DDelta = (\Delta_\i)_{\i \in \S_d}^\trans$, and $\W = [W_{\i,\j}]_{\i,~\j \in \S_d}$  analogously.
Letting $\nu_\j =   {d \choose \j } \int_{\RR^p} \u^{\j}\alpha(\u)\,\PP(\u)$ given a  kernel coefficient function $\alpha(\cdot)$, we solve the generalized eigenvalue problem in \eqref{eq:polyKDAeigen} for $\Nu=(\nu_\j)^\trans_{\j\in\S_d}$, and determine the population-level discriminant function as 
\begin{equation*}
f(\x) = \sum_{|\j|=d} \nu_\j\x^\j. 
\end{equation*}
Note that the size of $\DDelta$ and $\W$ is
$|\S_d|={d+p-1 \choose d}$, and while ordering of the indices in
$\S_d$ does not matter, the elements in $\DDelta$ and $\W$ should be
consistently indexed for specification of the eigenvalue problem.  
The following theorem summarizes the results so far.

\begin{theorem}
\label{thmhom}
Suppose that for each class, the distribution of $\X \in \RR^p$ has finite moments, 
$\EE_l [ \X^\i  ]$ and $\text{Cov}_l [\X^\i, \X^\j ]$ for all $\i,\j\in \S_d$. For the homogeneous polynomial kernel of degree $d$, $K_d(\x,\u)=(\x^\trans \u)^d$,
\begin{itemize}
\item[(i)] The kernel discriminant function maximizing the ratio of between-class variation relative to within-class variation is of the form
\begin{equation}
  f_d(\x) = \sum_{|\j|=d} \nu_\j \x^\j. \label{homdiscft}
\end{equation}

\item[(ii)] The coefficients, ${\Nu} = (\nu_{\i})^\trans_{\i \in \S_d}$, for the discriminant function satisfy the eigen-equation with $\lambda>0$:
  \begin{equation}
    \DDelta \DDelta^\trans  {\Nu} = \lambda \W{\Nu},  \label{eigenhomo}
  \end{equation}
  where $\DDelta = (\Delta_\i) ^\trans_{\i \in \S_d} $ is a vector of moment differences,
  $\Delta_\i = \EE_1[ \X^\i]- \EE_2[ \X^\i]$, and 
  $\W = [W_{\i,\j}]_{\i,~\j \in \S_d}$ is a matrix of pooled covariances,
  $W_{\i,\j} = \pi_1 \text{Cov}_1[\X^\i ,\X^\j]+\pi_2 \text{Cov}_2[\X^\i ,\X^\j]$.
 \end{itemize}
\end{theorem}

\bigskip

Alternatively, the discriminant function can be derived using an explicit feature map for the kernel. The expression of $K_d$ in \eqref{eq:polyK_d0} suggests
$\displaystyle \phi(\x) = \left( {d \choose \j}^{\frac{1}{2}} \x^\j\right) ^\trans_{\j \in \S_d}$
as a feature vector, and it can be shown that a direct application of LDA to 
the between-class and within-class variance matrices of $\phi(\X)$
leads to the same kernel discriminant. This result indicates that
employing homogeneous polynomial kernels in discriminant analysis has
the same effect as using the polynomial features of given degree in LDA.

\subsubsection{Inhomogeneous Polynomial Kernel}\label{InHomopolydp}

The inhomogeneous polynomial kernel of degree $d$ in $\RR^p$ can be expanded as  
\begin{equation*}
\tilde{K}_d(\x,\u) = (1+\x^\trans\u)^d = \sum_{m=0}^d {d \choose m} (\x^\trans\u)^m = \sum_{m=0}^d {d \choose m} \sum_{|\j|=m}{m \choose \j}\x^{\j}\u^{\j},  
\end{equation*}
which is a sum of all homogeneous polynomial kernels of degree up to $d$.
Since ${d \choose \j}={d \choose m}{m \choose \j}  $ for $\j \in \S_m$, $m=0,\ldots,d$, and the term with $m=0$ is 1, 
we can rewrite the kernel as 
\begin{equation*}
\tilde{K}_d(\x,\u) =  1+ \sum_{m=1}^d \sum_{|\j|=m} {d \choose \j} \x^{\j}\u^{\j}
= 1+ \sum_{|\j|=1}^d {d \choose \j} \x^{\j}\u^{\j}.
\end{equation*}
Note that $\sum_{m=1}^d \sum_{|\j|=m}$ is abbreviated to $\sum_{|\j|=1}^d$.
This kernel has the same structure as the homogenous polynomial kernel.
Using the relation, we can find the population kernel discriminant function similarly.
Recognizing that $\tilde{K}_d$ involves expanded polynomial features
in $p$ variables of total degree  0 to $d$: $1, \x, (\x^\j)_{|\j|=2},
\ldots, (\x^\j)_{|\j|=d}$, we define a vector of the mean differences of those features (excluding the constant $1$)
and a block matrix of their pooled covariances as follows:
\begin{equation*}
\tilde{\DDelta} = \left( \begin{array}{c}
        {\DDelta}_{1} \\ 
          \vdots   \\ 
        {\DDelta}_{d}
             \end{array}\right),
           \quad \mbox{ and } \quad
  \tilde{\W} = \left[ \begin{array}{ccc}
                     {\W}_{1,1}  &\ldots&{\W}_{1,d}\\
           \vdots &\ddots & \vdots \\ 
        {\W}_{d,1}&\ldots &{\W}_{d,d}
             \end{array}\right],              
\end{equation*} 
where  ${\DDelta}_{m} = (\Delta_{\i})^\trans_{\i \in \S_m}$  and
${\W}_{m,l} = [W_{\i,\j}]_{\i \in \S_m,\j \in \S_l}$ for all $m,~l = 1, \ldots,d$.  That is,  $\tilde{\DDelta}$ contains all the difference of the moments of degree 1 to $d$, and $\tilde{\W}$ has the covariances between all the monomials of degree 1 to $d$.
Thus, the size of the eigenvalue problem to solve becomes $\sum_{m=1}^d {m+p-1 \choose m} = {p+d \choose d}-1$.
The following theorem states similar results for the discriminant function with inhomogeneous polynomial kernel.

\begin{theorem}
\label{thminhom}
Suppose that for each class, the distribution of $\X \in \RR^p$ has finite moments, 
$\EE_l \left[ \X^{\i}  \right]$ and $\text{Cov}_l \left[\X^{\i}, \X^{\j} \right]$ for all $\i\in \S_m$, $\j\in \S_l$ and $m,l = 1,\ldots,d$. For the inhomogeneous polynomial kernel of degree $d$, $\tilde{K}_d(\x,\u)=(1+\x^\trans\u)^d$,
\begin{itemize}
\item[(i)] The kernel discriminant function maximizing the ratio of between-class variation relative to within-class variation is of the form
\begin{equation}
  \tilde{f}_d(\x) = \sum_{|\j|=1}^d   \tilde{\nu}_{\j}  \x^{\j}.  \label{inhomdiscft}
\end{equation}
    
\item[(ii)] The coefficients,
  $\tilde{\Nu}=(\tilde{\nu}_{\j})_{1\le |\j| \le d}^\trans$, for the discriminant function satisfy the eigen-equation with $\lambda>0$: 
 \begin{equation}
 \tilde{\DDelta} \tilde{\DDelta}^\trans \tilde{\Nu} = \lambda \tilde{\W}    \tilde{\Nu}.
 \label{eigeninhomo}
 \end{equation}

\end{itemize}
\end{theorem}

\subsection{Gaussian Discriminant}\label{subsec:GaussianKDA}

We extend the discriminant analysis with polynomial
kernels in the previous section to the Gaussian kernel.
For the extension, we use two representations for the Gaussian kernel: a
deterministic representation in Section~\ref{DeterG} and a randomized
feature representation in Section~\ref{FourierG}.

\subsubsection{Deterministic Representation of Gaussian Kernel}\label{DeterG}

We have seen so far that derivation of the population discriminant
function with polynomial kernels is aided by their expansion, or
equivalently, their explicit feature maps. Taking a similar approach to the 
Gaussian kernel, we could use the Maclaurin series of $e^x$ to express it as
\begin{eqnarray*}
  K_\omega(\x,\u) &=& \exp\left( -\frac{\|\x-\u\|^2}{2\omega^2}\right)
                      = \sum_{|\j|=0}^\infty \phi_{\j}(\x)\phi_{\j}(\u), 
\end{eqnarray*}
with
$\displaystyle \phi_{\j}(\x) =  \exp\left( -\frac{\|\x\|^2}{2\omega^2 }\right) \frac{1}{\sqrt{\j!}}\frac{\x^\j}{\omega^\j}$. 
While the structure of $K_\omega$ in this representation would permit 
similar derivations as before for the discriminant function, the result will 
depend on the expectations and covariances of $\phi_{\j}(\X)$ which
may not be easy to obtain analytically in general. 

Alternatively, we consider a representation of the kernel in the form 
that allows a direct use of polynomial features in much the same way as polynomial kernels.
We start with a one-dimensional case and then extend it to a multi-dimensional case. 
The one-dimensional Gaussian kernel with bandwidth $\omega$ can be written as 
\begin{equation}
  K_\omega(x,u)=\exp\left(-\frac{(x-u)^2}{2\omega^2}\right)
  = \exp\left( -\frac{x^2}{2\omega^2}\right)\sum_{m=0}^\infty He_m\left(\frac{x}{\omega}\right) \frac{u^m}{m!~\omega^m}.\label{monokernel}
\end{equation}
$He_m(x)$ are referred to as the probabilist's Hermite polynomials and defined as 
\begin{eqnarray*}
  He_m(x) &=& (-1)^m (\phi(x))^{-1}\frac{d^m}{dx^m}\phi(x),
\end{eqnarray*}
where $\phi$ is the density function of the standard normal distribution. 
The representation of $K_\omega$ in \eqref{monokernel} comes from the Hermite polynomial generating function: 
\begin{equation}
\exp\left( xu - \frac{1}{2}u^2\right) = \sum_{m=0}^\infty He_{m}(x) \frac{u^m}{m!}. \label{onedimgenft}
\end{equation}
It can be extended to a multivariate case using the vector-valued
Hermite polynomials introduced in \cite{holmquist1996d}.

For $\x \in \RR^p$ and $m\in \NN$,
the $p$-variate vector-valued Hermite polynomial of order $m$ is defined as
\begin{equation*}
 \H_{m}(\x) = (-1)^m(\Phi(\x))^{-1} \partial_\x^{\langle m \rangle}\Phi(\x),
\end{equation*}
where $\partial_\x^{\langle m \rangle} = \partial_\x \otimes
\partial_\x \otimes \cdots \otimes \partial_\x$ ($m$-times) is a
Kronecker product of the differential operator $\partial_\x  = (\frac{\partial}{\partial x_1},\ldots, \frac{\partial}{\partial x_p})^\trans$ and $\Phi$ is the product of $p$ univariate standard normal densities. Thus the components of $\H_{m}(\x)$ are a product of univariate Hermite polynomials whose total degree is $m$: $\H_{m}(\x) = (H_{\j}(\x))^\trans_{\j\in \S_m}$, where $H_{\j}(\x) = He_{j_1}(x_1)\cdots He_{j_p}(x_p)$ for each $\j \in \S_m$.

Using this notation, a multivariate version of the generating function \eqref{onedimgenft} can be written as
\begin{equation*}
\exp \left( \x^\trans\u-\frac{1}{2}\u^\trans\u \right) =
\sum_{m=0}^\infty \frac{1} {m!} \langle \H_m(\x), \u^{\langle m \rangle}\rangle,
\end{equation*} 
where $\u^{\langle m \rangle} = \u \otimes \u \otimes \cdots \otimes \u$ ($m$-times) and 
\begin{eqnarray*}
  \langle \H_m(\x), \u^{\langle m \rangle}\rangle
  = \sum_{\j \in \S_m} {m \choose \j}
      He_{j_1}(x_1)\cdots He_{j_p}(x_p)  u_1^{j_1}\cdots u_p^{j_p} 
  = \sum_{\j \in \S_m} {m \choose \j} H_\j(\x) \u^\j.
\end{eqnarray*}
Using the generating function for $\H_m$ and letting
$\x_\omega=(1/\omega)\x$ with bandwidth $\omega$, 
we get the following expansion for the multivariate Gaussian kernel: 
\begin{eqnarray*}
  K_\omega(\x, \u) &=&
  \exp \left( -\frac{\lVert \x-\u \rVert^2}{2 \omega^2}\right) 
 =\exp\left(-\frac{\lVert\x_\omega\rVert^2}{2}\right)\exp \left( \x_\omega^\trans\u_\omega-\frac{1}{2}\u^\trans_\omega \u_\omega  \right) \\\
&=& \exp\left(-\frac{\lVert\x_\omega\rVert^2}{2}\right)
    \sum_{|\j|=0}^\infty \frac{1}{\j!} H_{\j}(\x_\omega)
    \u_\omega^{\j}. 
\end{eqnarray*}
Further with the definition of 
$ \displaystyle 
\tilde{H}_{\j}(\x_\omega)=\frac{1}{\j!
    \omega^\j}\exp\left(-\frac{\lVert\x_\omega\rVert^2}{2}\right)H_{\j}(\x_\omega)$,
 the kernel is represented as 
 \begin{equation} 
   K_\omega(\x, \u)=\sum_{|\j|=0}^\infty \tilde{H}_{\j}(\x_\omega) \u^{\j}.    \label{gaussianrep}
 \end{equation}
 Although this representation is asymmetric in $\x$ and $\u$,
 it facilitates similar derivations of the
 generalized eigenvalue problem and population kernel discriminant as
 with polynomial kernels, but using the entirety of polynomial features.
 
With this representation, it is easy to show that
\begin{eqnarray*}
  \EE_1\left[ K_\omega(\x,\X)\right] - \EE_2[ K_\omega(\x,\X)]
  &=& \sum_{|\j|=0}^\infty \tilde{H}_{\j}(\x_\omega) \left\{ \EE_1[ \X^{\j} ]-\EE_2[ \X^{\j}]\right\} \\
  &=& \sum_{|\j|=0}^\infty \tilde{H}_{\j}(\x_\omega) \Delta_\j
      =\sum_{|\j|=1}^\infty \tilde{H}_{\j}(\x_\omega) \Delta_\j,
\end{eqnarray*}
which involves the moments of the distribution rather than the
expectations of $\tilde{H}_{\j}(\X_\omega)$.
Note that the last equality is due to $\Delta_{\mathbf{0}}=0$ for $\X^{\mathbf{0}}=1$. 
Thus the between-class variation function is given as
\begin{eqnarray*}
B_K(\x,\u) &=&\sum_{|\i|=1}^\infty \sum_{|\j|=1}^\infty \Delta_{\i}\Delta_{\j}\tilde{H}_{\i}(\x_\omega) \tilde{H}_{\j}(\u_\omega).
\end{eqnarray*}
Similarly the within-class variation function is given as
\begin{eqnarray*}
W_K(\x,\u) &=&\sum_{|\i|=1}^\infty \sum_{|\j|=1}^\infty  W_{\i,\j}\tilde{H}_{\i}(\x_\omega) \tilde{H}_{\j}(\u_\omega).
\end{eqnarray*}
Therefore, the eigenvalue problem in \eqref{eq:kldafcn} with the 
Gaussian kernel is given by
\begin{equation}
\sum_{|\i|=1}^\infty \sum_{|\j|=1}^\infty   \Delta_{\i} \Delta_{\j}\nu_{\j}\tilde{H}_{\i}(\x_\omega)
  = \lambda \sum_{|\i|=1}^\infty \sum_{|\j|=1}^\infty
  \W_{\i,\j}\nu_{\j}\tilde{H}_{\i}(\x_\omega),
  \label{GaussianEigen}
\end{equation}
where $\nu_{\j} = \int_\Xcal \tilde{H}_{\j}(\u_\omega)\; \alpha(\u) \; d\PP(\u)$.

To find $\nu_\j$ satisfying \eqref{GaussianEigen} for every $\x_\omega$,
the coefficients of $ \tilde{H}_\i(\x_\omega)$ on both sides must 
equal for all $\i \in \S_m$,  $m\in \NN$. This entails the following
system of an infinite number of linear equations for $\nu_{\j}$:
\begin{align}
  \Delta_{\i} \sum_{|\j|=1}^\infty  \Delta_{\j} \nu_{\j}
                 &= \lambda     \sum_{|\j|=1}^\infty W_{\i,\j}\nu_{\j},
                       \quad   \i \in\S_m,  ~ m\in \NN,  \label{infinitedimeigen1}
\end{align}
and the resulting discriminant function of the form:
$f(\x)=\sum_{|\j|=1}^\infty \nu_\j\x^\j$.

For a finite dimensional approximation of the population discriminant
function, we may consider truncation of the kernel representation in \eqref{gaussianrep} at $|\j|=N$:
\begin{eqnarray*}
  K_N(\x,\u)&=& \sum_{|\j|=0}^N  \tilde{H}_{\j}(\x_\omega) \u^{\j}.
\end{eqnarray*}
This approximation brings the corresponding truncation of the system
of linear equations for the generalized eigenvalue problem in
\eqref{infinitedimeigen1}. As a result, the eigenvalue equation 
coincides with that for the inhomogeneous polynomial kernel of degree
$N$ in Theorem~\ref{thminhom}, and so does the truncated
discriminant function.
As more polynomial features are added or $N$ increases, the largest eigenvalue satisfying equation \eqref{eigeninhomo} increases. 
Adding subscript $N$ to $\lambda$, $\tilde{\DDelta}$ and $\tilde{\W}$
to indicate the degree clearly, let $\displaystyle \lambda_N=\max_{\Nu} \frac{\Nu^\trans \tilde{\DDelta}_{N}\tilde{\DDelta}_{N}^\trans \Nu }{\Nu^\trans   \tilde{\W}_N\Nu}$. 
The moment difference vector $\tilde{\DDelta}_{N}$ and the within-class covariance matrix $\tilde{\W}_{N}$ expand with $N$, including all the elements up to degree $N$.
This nesting structure produces an increasing sequence of $\lambda_N$.
It is because maximization of the ratio for degree $N$ amounts to that
for degree $N+1$ with a limited space for $\Nu$.
In Section~\ref{subsec:simulation},
we will study the relation between polynomial and Gaussian 
discriminants numerically under various scenarios and discuss the effect of 
$N$ on the quality of the discriminant function.

\subsubsection{Fourier Feature Representation of Gaussian kernel}\label{FourierG}

In addition to the polynomial approximation presented in the
previous section, a stochastic approximation to the Gaussian kernel can be
used for population analysis.
\cite{rahimi2008random} examined approximation of shift-invariant
kernels in general using random Fourier features for fast large-scale optimization
with kernels. They proposed 
the following representation for the Gaussian kernel using random 
features of the form $z_{\w}(\x)=(\cos(\w^\trans \x),\sin(\w^\trans \x))^\trans$:
\begin{equation}
  K_\omega(\x,\u) =\EE_{\w}\left[z_{\w}(\x)^\trans  z_{\w}(\u)\right]
  \label{Kzapp},
\end{equation}
where $\w$ is a random vector from a multivariate normal distribution with mean zero
and covariance matrix $\frac{1}{\omega^2}I_p$.
This representation comes from Bochner's theorem \citep{rudin2017fourier}, which
 describes the correspondence between a positive
definite shift-invariant kernel and the Fourier transform of a
nonnegative measure. 
The feature map $z_{\w}(\cdot)$ projects $\x$ onto a random direction
$\w$ first and then takes sinusoidal transforms. Their frequency
depends on the norm of $\w$. A large bandwidth $\omega$ for the
Gaussian kernel implies realization of $\w$ with a small norm on average,
which  generally entails a low frequency for the sinusoids.

The representation in \eqref{Kzapp} suggests a Monte Carlo approximation of the
kernel. Suppose that $\w_i$, $i=1,\ldots, D$ are randomly generated
from $N_p(\mathbf{0}, \frac{1}{\omega^2}I_p)$.  Defining random Fourier features $z_{\w}(\x)$
with $\w=\w_i$, we can approximate the Gaussian kernel using a sample
average as follows:
\begin{equation*}
  K_\omega(\x,\u)=\exp\left( -\frac{\|\x-\u\|^2}{2\omega^2}\right)
  \approx  \frac{1}{D} \sum_{i=1}^D z_{\w_i}(\x)^\trans z_{\w_i}(\u).   
\end{equation*}
This average can be taken as an unbiased estimate of the kernel, and
its precision is controlled by $D$.
Concatenating these $D$ random components $z_{\w_i}(\x)$, we can also
see that the stochastic approximation above amounts to defining
\begin{equation*}
  \vec{\Z}_{D}(\x) =  \frac{1}{\sqrt{D}}(z_{\w_1}(\x)^\trans,\ldots,z_{\w_D}(\x)^\trans)^\trans
\end{equation*} 
as a randomized feature map for the kernel.

Using the random Fourier features, 
we approximate  the between-class variation function $B_K(\x,\u)$
and within-class variation function $W_K(\x,\u)$ as follows:
 \begin{eqnarray*}
B_K(\x,\u) &\approx&  \frac{1}{D^2} \sum_{i=1}^D\sum_{j=1}^D
                     z_{\w_i}(\x)^\trans \Delta_{\w_i}\Delta_{\w_j}^\trans z_{\w_j}(\u)\\
W_K(\x,\u) &\approx& \frac{1}{D^2} \sum_{i=1}^D
                     \sum_{j=1}^D  z_{\w_i}(\x)^\trans W_{\w_i,\w_j} z_{\w_j}(\u),\\
 \end{eqnarray*}
 where $\Delta_{\w_i}=\EE_1\left[z_{\w_i}(\X)\right] -
 \EE_2\left[z_{\w_i}(\X)\right]$ and 
$W_{\w_i,\w_j}=\pi_1\text{Cov}_1 \left[
  z_{\w_i}(\X),z_{\w_j}(\X)\right] +\pi_2\text{Cov}_2 \left[
  z_{\w_i}(\X),z_{\w_j}(\X)\right]$.
 Then we can define a randomized version of the eigenvalue problem in
\eqref{eq:kldafcn} with these approximations.
Let $\hat{\alpha}(\cdot)$ denote the solution to the problem with $\lambda>0$
and define $\Nu_i=\int z_{\w_i}(\u)\hat{\alpha}(\u) \, d\PP(\u)$.
Similar arguments as before lead to the following generalized eigenvalue
problem to determine $\Nu=(\Nu_i^\trans)^\trans$:
\begin{equation*} 
 \hat{\DDelta} \hat{\DDelta} ^\trans \Nu=\lambda \hat{\W} \Nu,
\end{equation*}
where $\hat{\DDelta} =(\Delta_{\w_i}^\trans)^\trans$ and
$\hat{\W}=[W_{\w_i,\w_j}]$ for $i, j=1,\ldots,D$.
Given $\Nu$, the approximate Gaussian discriminant obtained via random Fourier features is
\begin{equation}
  f_D(\x) = \frac{1}{D}\sum_{i=1}^D \Nu_i^\trans z_{\w_i}(\x).
  \label{RFFdisc}
 \end{equation}

 Rather than sine and cosine pairs, we could also use phase-shifted
 cosine features only to approximate the Gaussian kernel as suggested
 in \cite{rahimi2008random}  and \cite{Rahimi:Recht2008uniform}.
 Let $z_{\w,b}(\x)=\sqrt{2}\cos(\w^\trans \x+b)$ with 
an additional phase parameter $b$ which is independent of $\w$ and
distributed uniformly on $(0,2\pi)$. Then using a
trigonometric identity, we can verify that
\[K_\omega(\x,\u)=\EE_{\w,b}\left[ z_{\w,b}(\x) z_{\w,b}(\u)\right]
  =\EE_{\w,b}\left[ 2 \cos(\w^\trans \x+b)\cos(\w^\trans\u+b)\right].\]
Given $\w$ and $b$, if $\X$ is distributed with $N_p(\Mu, \Sigma)$, we
can show that
\[\EE_{\X}\left[\cos(\w^\trans \X+b)\right]
=\exp(-\frac{1}{2}\w^\trans \Sigma \w)\cos(\w^\trans\Mu+b).\]
Thus in
the classical LDA setting of $\PP_j=N(\Mu_j, \Sigma)$ for $j=1,2$,  this
Fourier feature lets us focus on the difference in
$\cos(\w^\trans\Mu_j+b)$ rather than $\Mu_j$.


\section{Numerical Studies}\label{sec:simstudy}

This section illustrates the relation between the data distribution and kernel discriminants discussed so far through simulation studies and an application to real data.

\subsection{Simulation Study}\label{subsec:simulation}

\begin{figure}[bth!]
\centering
\begin{subfigure}{.4\linewidth}
 \includegraphics[width=1\linewidth]{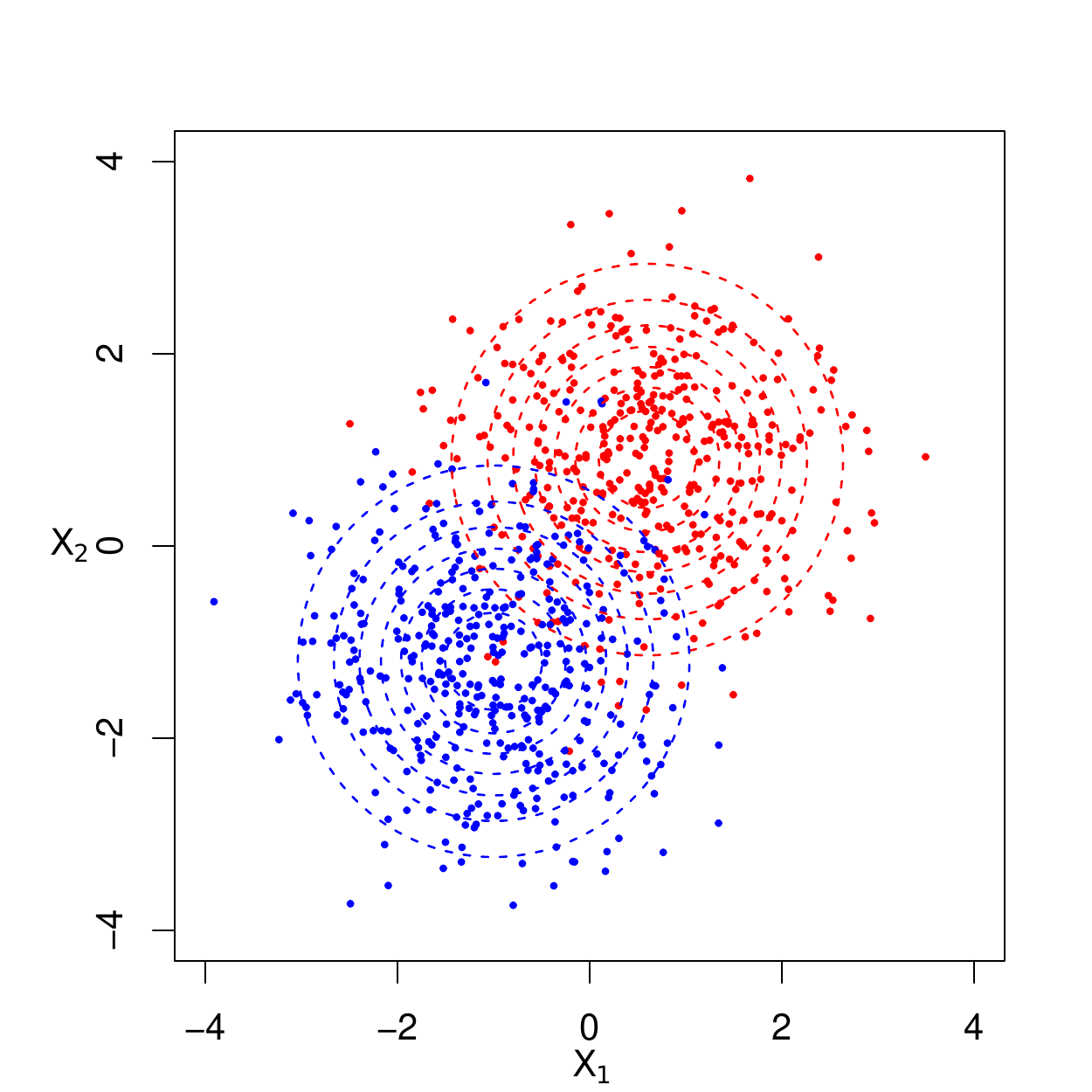}
 \caption{Scenario 1}
\end{subfigure}
\begin{subfigure}{.4\linewidth}
 \includegraphics[width=1\linewidth]{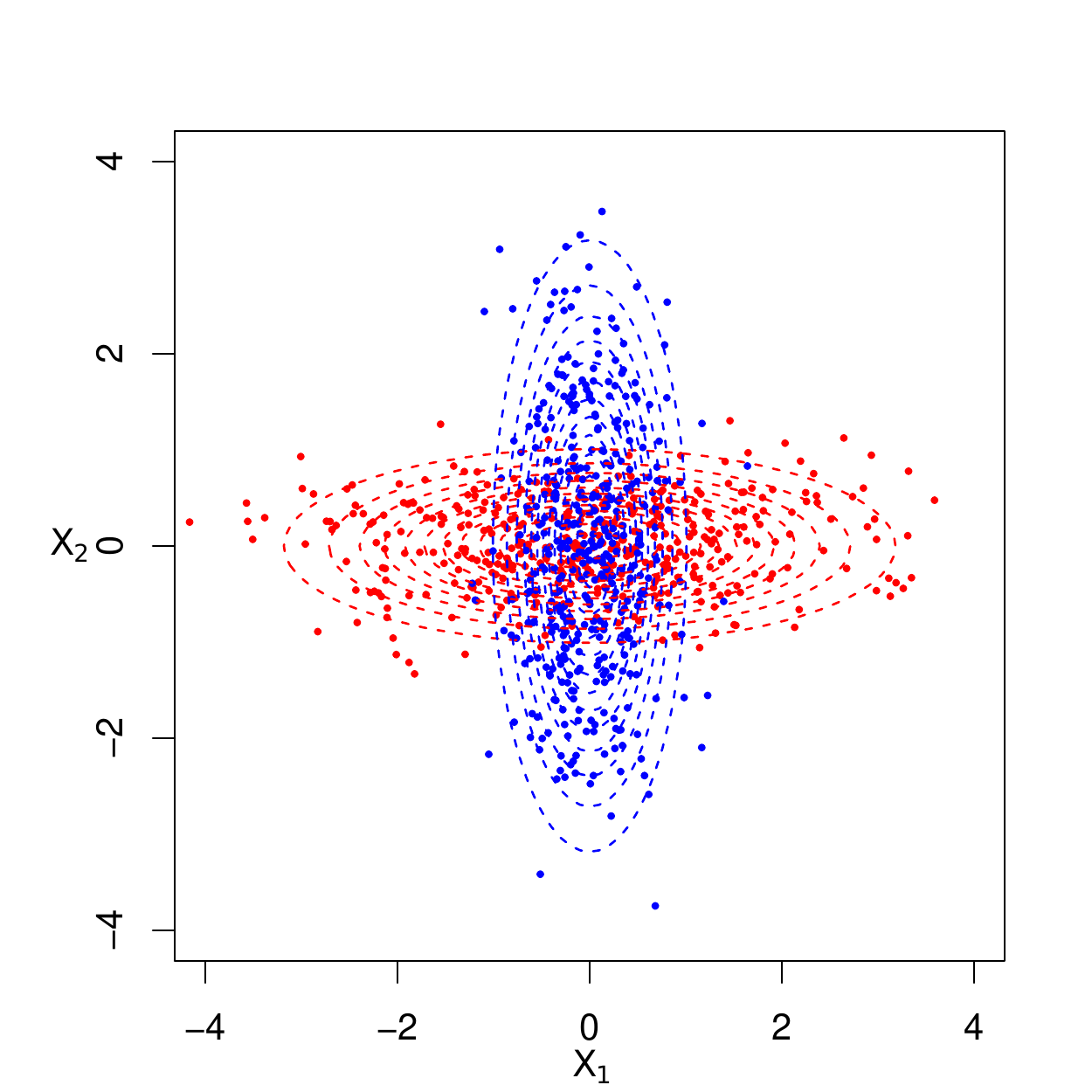}
  \caption{Scenario 2}
\end{subfigure}

\caption{Scatterplots of the samples simulated from a mixture of two normal distributions with contours of the probability densities for each class overlaid in two settings: (a) Scenario 1 and (b) Scenario 2.
}
\label{Scenarios}
\end{figure}

We numerically study the population discriminant functions
in \eqref{homdiscft}, \eqref{inhomdiscft}, and \eqref{RFFdisc}  
with both polynomial and Gaussian kernels, and examine their
relationship with the underlying data distributions for two
classes. For illustration, we consider two scenarios where each class
follows a bivariate normal distribution. In Scenario 1, two classes
have different means ($\Mu_1=(0.6, 0.9)^\trans$ and
$\Mu_2=(-1.0,-1.2)^\trans$) but the same covariance
($\Sigma_1=\Sigma_2=I_2$), and in Scenario 2, they have the same mean
($\Mu_1=\Mu_2=\mathbf{0}$) but different covariances
($\Sigma_1=\mbox{diag}(2, 0.2)$ and $\Sigma_2=\mbox{diag}(0.2, 2)$).
Figure~\ref{Scenarios} shows the scatter plots of samples generated
from each scenario with 400 data points in each class (red: class 1
and blue: class 2) under the assumption that two classes are equally likely. 

\subsubsection{Polynomial Kernel}\label{theorypolykernel}

Under each scenario, we find the population discriminant functions in \eqref{homdiscft} and \eqref{inhomdiscft} with polynomial kernels of degree 1 to 4 and examine the effect of the degree on the discriminants.
To determine $f_d(\x)$, we first obtain the population moment
differences $\DDelta$ and covariances $\W$ explicitly and solve the
eigenvalue problem in \eqref{eigenhomo}. Similarly we determine
$\tilde{f}_d(\x)$ with $\tilde{\DDelta}$ and  $\tilde{\W}$.
Tables~\ref{coefftableS1} and~\ref{coefftableS2} present
the coefficients for the polynomial discriminants $f_d(\x)$ and
$\tilde{f}_d(\x)$ in each scenario, which are the solution $\Nu$ or  $\tilde{\Nu}$
(eigenvector) normalized to unit length.

\begin{table}[ht!]
 \caption{Coefficients for the population polynomial discriminants under Scenario 1. 
}
\label{coefftableS1}
\centering
\begin{tabular}{c|cccc|ccrr}
\hline\hline
\multicolumn{1}{l|}{} & \multicolumn{4}{c|}{Homogeneous polynomial}                                                                                 & \multicolumn{4}{c}{Inhomogeneous polynomial}                                                                                                                                                       \\
{Term}             & {$f_{\1}(\x)$} & {$f_{\2}(\x)$} & {$f_{\3}(\x)$} & {$f_{\4}(\x)$} & {$\tilde{f}_{\1}(\x)$} & {$\tilde{f}_{\2}(\x)$} & {$\tilde{f}_{\3}(\x)$} &{$\tilde{f}_{\4}(\x)$} \\ [3pt]\hline
$x_1$                              & \multicolumn{1}{r}{0.6060}    & -                             & -                             & -                             & \multicolumn{1}{r}{0.6060}            & \multicolumn{1}{r}{0.6060}            & 0.6033                                                    & 0.6033                                                    \\
$x_2$                              & \multicolumn{1}{r}{0.7954}    & -                             & -                             & -                             & \multicolumn{1}{r}{0.7954}            & \multicolumn{1}{r}{0.7954}            & 0.7919                                                    & 0.7919                                                    \\ \cline{1-6}
$x_1^2$                            & -                             & \multicolumn{1}{r}{-0.4461}   & -                             & -                             & -                                     & \multicolumn{1}{r}{0.0000}            & -0.0141                                                   & -0.0141                                                   \\
$x_1x_2$                           & -                             & \multicolumn{1}{r}{-0.8376}   & -                             & -                             & -                                     & \multicolumn{1}{r}{0.0000}            & -0.0369                                                   & -0.0369                                                   \\
$x_2^2$                            & -                             & \multicolumn{1}{r}{-0.3154}   & -                             & -                             & -                                     & \multicolumn{1}{r}{0.0000}            & -0.0242                                                   & -0.0242                                                   \\ \cline{1-7}
$x_1^3$                            & -                             & -                             & \multicolumn{1}{r}{0.6412}    & -                             & -                                     & -                                     & -0.0118                                                   & -0.0118                                                   \\
$x_1^2x_2$                         & -                             & -                             & \multicolumn{1}{r}{0.3105}    & -                             & -                                     & -                                     & -0.0465                                                   & -0.0465                                                   \\
$x_1x_2^2$                         & -                             & -                             & \multicolumn{1}{r}{-0.2277}   & -                             & -                                     & -                                     & -0.0610                                                   & -0.0610                                                   \\
$x_2^3$                            & -                             & -                             & \multicolumn{1}{r}{0.6637}    & -                             & -                                     & -                                     & -0.0267                                                   & -0.0267                                                   \\ \cline{1-8}
$x_1^4$                            & -                             & -                             & -                             & \multicolumn{1}{r|}{-0.2575}  & -                                     & -                                     & \multicolumn{1}{c}{-}                                     & 0.0000                                                    \\
$x_1^3x_2$                         & -                             & -                             & -                             & \multicolumn{1}{r|}{-0.6186}  & -                                     & -                                     & \multicolumn{1}{c}{-}                                     & 0.0000                                                    \\
$x_1^2x_2^2$                       & -                             & -                             & -                             & \multicolumn{1}{r|}{0.3860}   & -                                     & -                                     & \multicolumn{1}{c}{-}                                     & 0.0000                                                    \\
$x_1x_2^3$                         & -                             & -                             & -                             & \multicolumn{1}{r|}{-0.6146}  & -                                     & -                                     & \multicolumn{1}{c}{-}                                     & 0.0000                                                    \\
$x_2^4$                            & -                             & -                             & -                             & \multicolumn{1}{r|}{-0.1563}  & -                                     & -                                     & \multicolumn{1}{c}{-}                                     & 0.0000                \\\hline                                   
\end{tabular}
\end{table}

\noindent {\bf Scenario 1:}
Fisher's linear discriminant analysis is optimal in this scenario. Since the common covariance matrix is $I_2$, the linear discriminant is simply determined by the direction of the mean difference, which is $\Mu_1-\Mu_2 = (1.6,2.1)^\trans$. This gives $f^*(\x)=1.6x_1 + 2.1x_2$ as an optimal linear discriminant defined up to a multiplicative constant. 
From Table~\ref{coefftableS1}, we first notice that the coefficient
vector for the population linear discriminant,  $f_1(\x)$, $\Nu =
(0.6060, 0.7954)^\trans$, is a normalized mean difference. Further we
observe that the coefficients for the discriminants with inhomogeneous
polynomial kernels, $\tilde{f}_{\1}(\x)$ and $\tilde{f}_{\2}(\x)$, are
also proportional to the mean difference.

\begin{figure}[th!]
\centering
\includegraphics[width=1\linewidth]{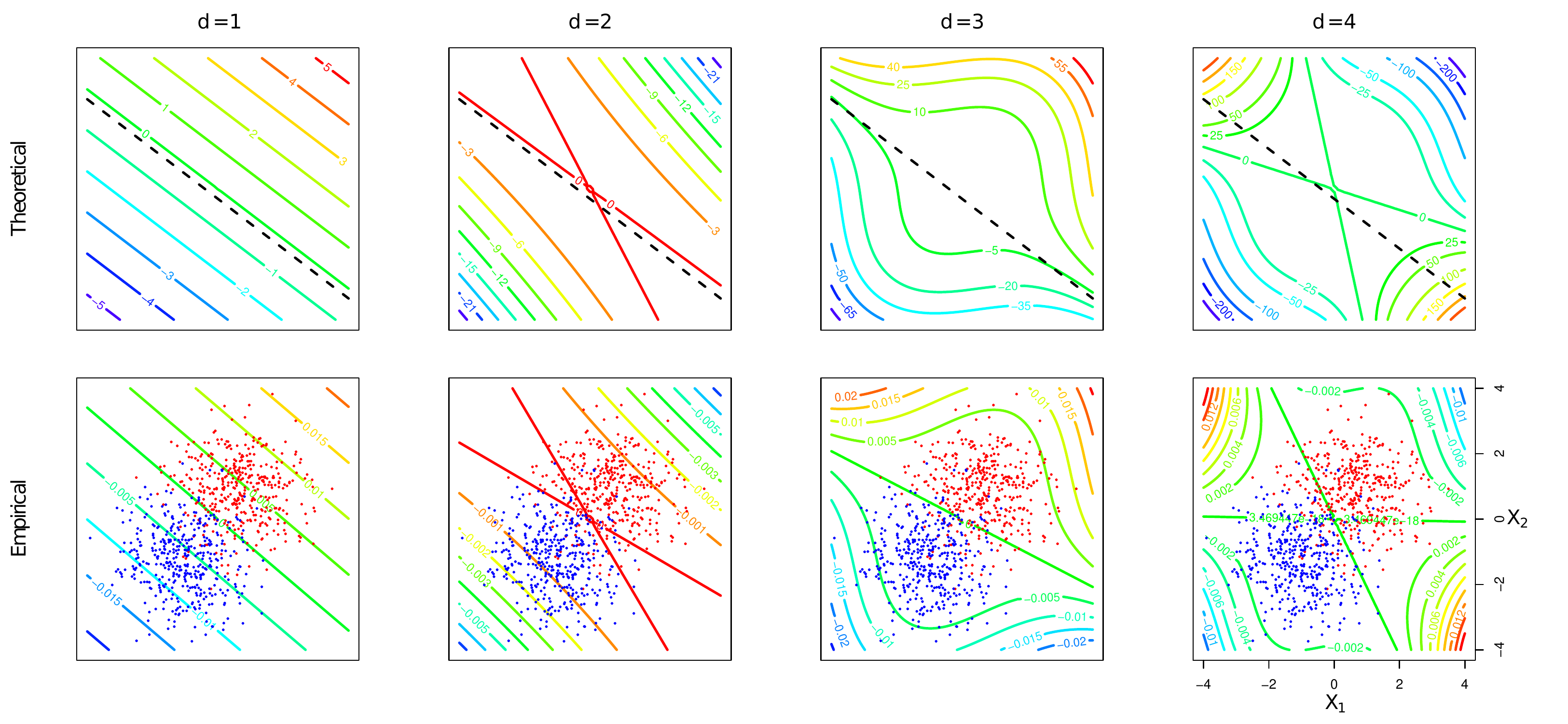}
\caption{Contours of the population discriminant functions with homogeneous polynomial kernels of degree 1 to 4 (upper panels from left to right) and their corresponding sample counterparts (lower panels) under Scenario 1. The black dashed lines are the optimal classification boundary.}
\label{theodiscpolyhomoS1}
\end{figure}

\begin{figure}[ht!]
\centering
\includegraphics[width=.8\linewidth]{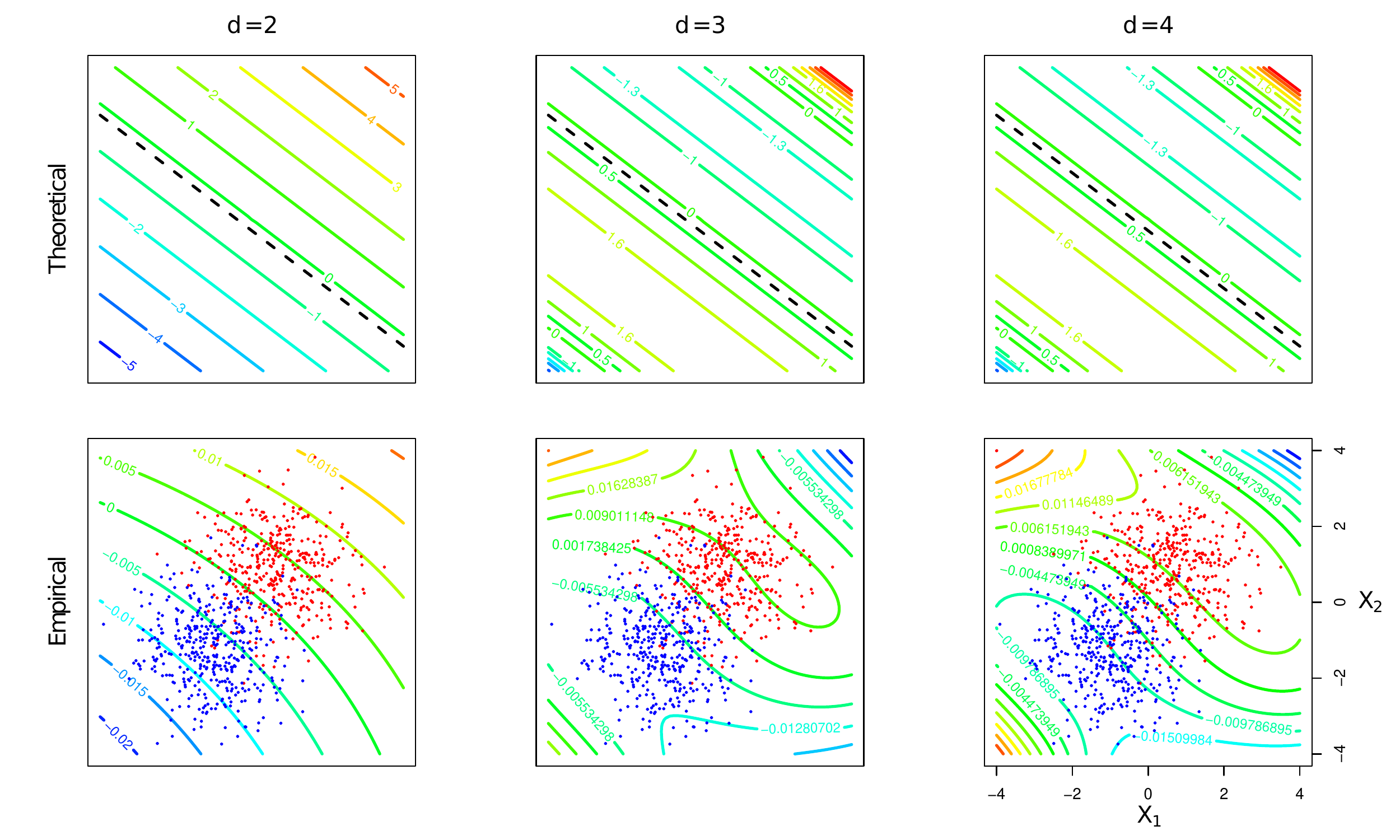}
\caption{Contours of the population discriminant functions with inhomogeneous polynomial kernels of degree 2 to 4 (upper panels from left to right) and their corresponding sample counterparts (lower panels) under Scenario 1. The black dashed lines are the optimal classification boundary.}
\label{theodiscpolyinhomoS1}
\end{figure}

Figures~\ref{theodiscpolyhomoS1}
and~\ref{theodiscpolyinhomoS1} display the polynomial discriminants
identified in Table~\ref{coefftableS1}.
The first row of Figure~\ref{theodiscpolyhomoS1} shows contours of the population discriminants  with homogenous polynomial kernels.
High to low discriminant scores correspond to red to blue contours. The black dashed line is $1.6x_1 + 2.1x_2 = 0.635$, which is the classification boundary from Fisher's linear discriminant analysis. The second row of Figure~\ref{theodiscpolyhomoS1} presents  the corresponding sample embeddings obtained by performing a kernel discriminant analysis to the given samples.  Figure~\ref{theodiscpolyinhomoS1} shows contours of  both versions with inhomogeneous polynomial kernels of degree 2 to 4, omitting degree 1 as they are identical to those with the linear kernel in Figure~\ref{theodiscpolyhomoS1}.

The population discriminants and sample versions are similar in terms of shape and direction of change in contours. With odd-degree homogeneous polynomial kernels, we observe that the contours change in the direction of the mean difference, indicating that odd degrees are effective in this setting.
The even-degree discriminants, however, are of hyperbolic paraboloid
shape, varying in a way that masks the class difference completely. By
contrast, the degree doesn't affect the major direction of change in
the population discriminants with inhomogeneous polynomial
kernels. Their  variation seems to occur only in the direction of the
mean difference. Table~\ref{coefftableS1} confirms that the resulting 
discriminants $\tilde{f}_d(\x)$ are identical for degrees $d=2k-1$ and $2k$, $k=1,2$.

\begin{table}[th!]
\caption{Coefficients for the population polynomial discriminants under Scenario 2.}\label{coefftableS2} 
\centering
\begin{tabular}{c|cccc|ccrr}
\hline\hline
\multicolumn{1}{c|}{}  & \multicolumn{4}{c|}{Homogeneous polynomial} &\multicolumn{4}{c}{Inhomogeneous polynomial}                                         \\ 
{Term} & {$f_{\1}(\x)$}      & {$f_{\2}(\x)$}    & {$f_{\3}(\x)$}           & {$f_{\4}(\x)$}           &{$\tilde{f}_{\1}(\x)$}       & {$\tilde{f}_{\2}(\x)$}      & $\tilde{f}_{\3}(\x)$ & $\tilde{f}_{\4}(\x)$ \\[3pt]\hline
$x_1$                  & \multicolumn{1}{r}{0.00} & -                          & -                           & -                            & \multicolumn{1}{r}{0.00} & \multicolumn{1}{r}{0.0000} & 0.0000                                 & 0.0000                                 \\
$x_2$                  & \multicolumn{1}{r}{0.00} & -                          & -                           & -                            & \multicolumn{1}{r}{0.00} & \multicolumn{1}{r}{0.0000} & 0.0000                              & 0.0000                                 \\ \cline{1-6}
$x_1^2$                & -                       & \multicolumn{1}{r}{0.7071} & -                           & -                            & -                       & \multicolumn{1}{r}{0.7071} & 0.7071                               & 0.7063                              \\
$x_1x_2$               & -                       & \multicolumn{1}{r}{0.0000} & -                           & -                            & -                       & \multicolumn{1}{r}{0.0000} & 0.0000                                & 0.0000                               \\
$x_2^2$                & -                       & \multicolumn{1}{r}{-0.7071} & -                           & -                            & -                       & \multicolumn{1}{r}{-0.7071} & -0.7071                              & -0.7063                               \\ \cline{1-7}
$x_1^3$                & -                       & -                          & \multicolumn{1}{r}{0.0000}  & -                            & -                       & -                       & 0.0000                                & 0.0000                                \\
$x_1^2x_2$             & -                       & -                          & \multicolumn{1}{r}{0.0000}  & -                            & -                       & -                       & 0.0000                                & 0.0000                                \\
$x_1x_2^2$             & -                       & -                          & \multicolumn{1}{r}{0.0000} & -                            & -                       & -                       & 0.0000                                & 0.0000                               \\
$x_2^3$                & -                       & -                          & \multicolumn{1}{r}{0.0000}  & -                            & -                       & -                       & 0.0000                                & 0.0000                                \\ \cline{1-8}
$x_1^4$                & -                       & -                          & -                           & \multicolumn{1}{r|}{0.7071}  & -                       & -                       & \multicolumn{1}{c}{-}                  & -0.0335                                     \\
$x_1^3x_2$             & -                       & -                          & -                           & \multicolumn{1}{r|}{0.0000}  & -                       & -                       & \multicolumn{1}{c}{-}                  & 0.0000                                      \\
$x_1^2x_2^2$           & -                       & -                          & -                           & \multicolumn{1}{r|}{0.0000} & -                       & -                       & \multicolumn{1}{c}{-}                  & 0.0000                                      \\
$x_1x_2^3$             & -                       & -                          & -                           & \multicolumn{1}{r|}{0.0000}  & -                       & -                       & \multicolumn{1}{c}{-}                  & 0.0000                                      \\
$x_2^4$                & -                       & -                          & -                           & \multicolumn{1}{r|}{-0.7071}  & -                       & -                       & \multicolumn{1}{c}{-}                  &0.0335        \\\hline                       
\end{tabular}
\end{table}

\begin{figure}[tbh!]
\centering
\includegraphics[width=1\linewidth]{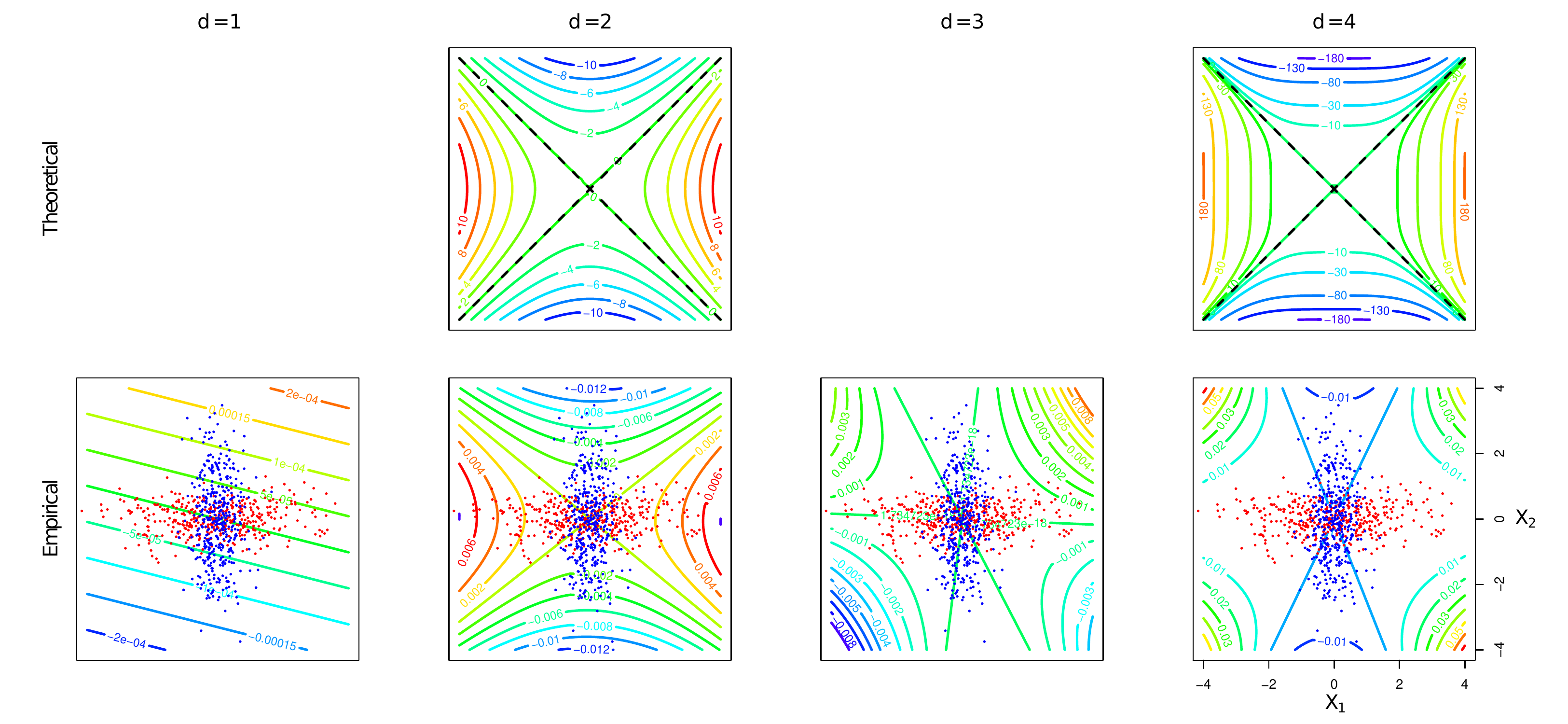} 
\caption{Contours of the population discriminant functions with
  homogeneous polynomial kernels of degree 1 to 4 (upper panels) and
  their sample counterparts (lower panels) under Scenario 2. The black dashed lines are the optimal classification boundaries.  }
\label{theodiscpolyhomoS2}
\end{figure}

\begin{figure}[tbh!]
\centering
\includegraphics[width=.8\linewidth]{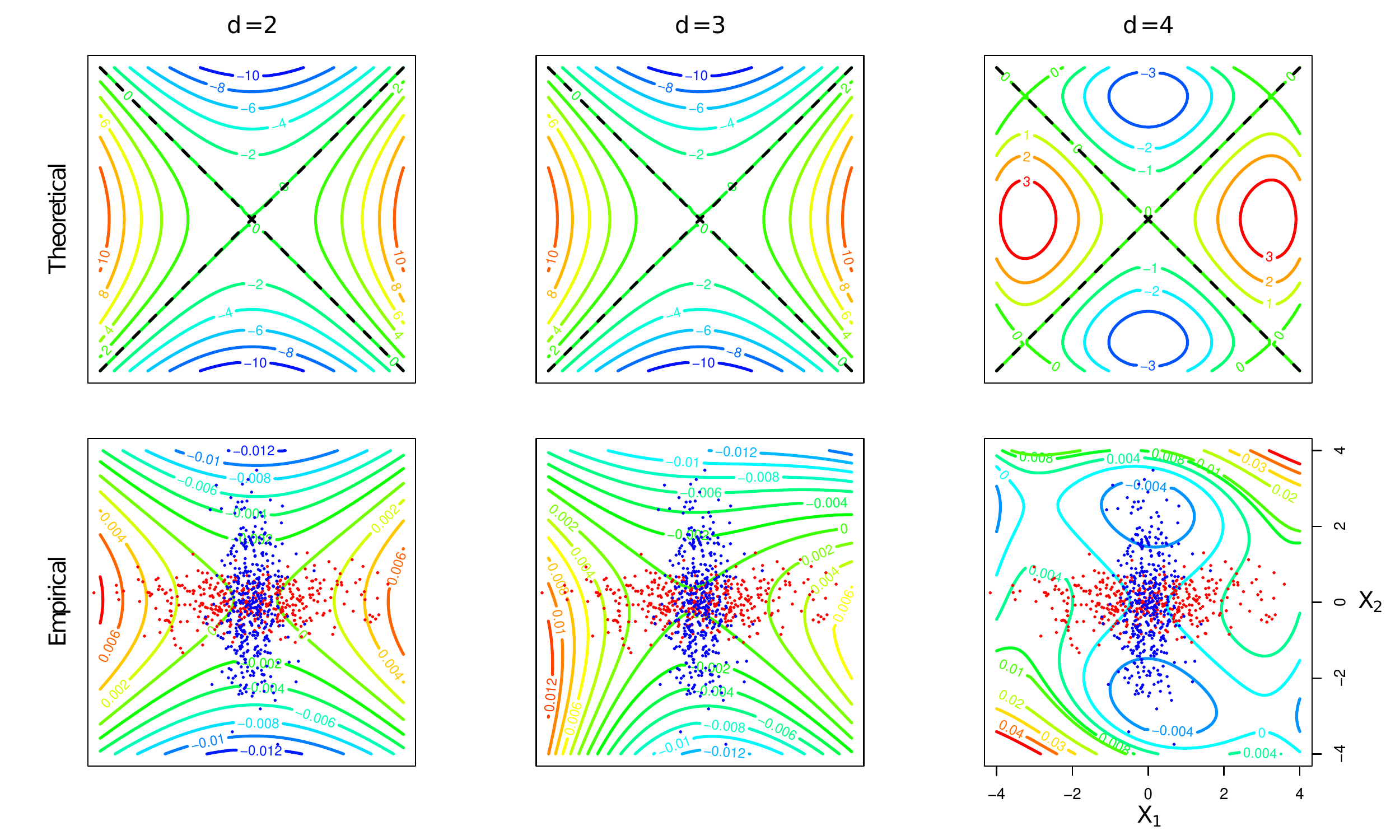} 
\caption{Contours of the population discriminant functions with
  inhomogeneous polynomial kernels of degree 2 to 4 (upper panels) and
  their sample counterparts (lower panels) under Scenario 2. The black dashed lines are the optimal classification boundaries. }
\label{theodiscpolyinhomoS2}
\end{figure}

\noindent {\bf Scenario 2:}
In this scenario,  using the true densities, the optimal decision boundary is found to be $(x_1 + x_2)(x_1 - x_2)=0$, and the optimal discriminant function is $f^*(\x)=x_1^2-x_2^2$, which is a homogeneous polynomial of degree 2. In contrast with Scenario 1, even-degree features are discriminative in this setting. 
Note that the  coefficients of $f_{\2}(\x)$, $\tilde{f}_{\2}(\x)$ and $\tilde{f}_{\3}(\x)$ in Table~\ref{coefftableS2} are proportional to those of $f^*(\x)$.   Odd-degree homogeneous polynomials produce a degenerate discriminant in this setting.
The quadratic discriminant, $f_{\2}(\x)= 0.7071x_1^2 - 0.7071x_2^2$, is a normalized version of  $f^*(\x)$. With degree 4 homogeneous polynomial kernel, we have $f_{\4}(\x)=0.7071x_1^4 - 0.7071x_2^4$, which has the optimal discriminant as its factor. Contours of these polynomial discriminants are displayed in the first row of Figure~\ref{theodiscpolyhomoS2}. The black dashed lines are the optimal decision boundaries. The second row of Figure~\ref{theodiscpolyhomoS2} presents the corresponding nonlinear kernel embeddings of degree 1 to 4 induced by the samples. Figure~\ref{theodiscpolyinhomoS2} shows contours of both versions (theoretical in the first row and empirical in the second row) with inhomogeneous polynomial kernels of degree 2 to 4, omitting the degenerate linear case in Table~\ref{coefftableS2}.

Similar to Scenario 1, we observe that the population discriminant functions and their sample counterparts in Figures~\ref{theodiscpolyhomoS2} and~\ref{theodiscpolyinhomoS2} exhibit similarity in terms of shape and direction of change in contours.  The contours of the population quadratic and quartic discriminants in Figure~\ref{theodiscpolyhomoS2} show symmetry along each variable axis. Quadratic features contain all information necessary for discrimination in this scenario. Even-degree features successfully discriminate the two classes while odd-degree features completely fail as shown in Figure~\ref{theodiscpolyhomoS2}. Nonlinear inhomogeneous polynomial kernels with even-degree features enable proper classification as illustrated in Figure~\ref{theodiscpolyinhomoS2}.
Inhomogeneous polynomial kernels of degree $2k+1$ and $2k$ produce identical discriminants in this setting.

\subsubsection{Gaussian Kernel}\label{theoryGkernel}

We examine Gaussian discriminant functions under each scenario using
two types of approximation to the Gaussian kernel
discussed earlier.\\

\noindent{\bf Deterministic representation:}
Truncation of the deterministic representation of the Gaussian kernel at a certain degree 
leads to the population polynomial discriminant using the inhomogeneous polynomial kernel  of the same degree. Thus to approximate the population Gaussian discriminant, we need to choose an appropriate degree for truncation.
As the truncation degree $N$ increases, the largest (and only nonzero) eigenvalue $\lambda_N$ as a measure of class separation naturally increases. We may stop at $N$ where the increment in $\lambda_N$ is negligible. 

\begin{figure}[ht!]
\centering
\begin{subfigure}{.45\linewidth}
\centering
 \includegraphics[width=1\linewidth]{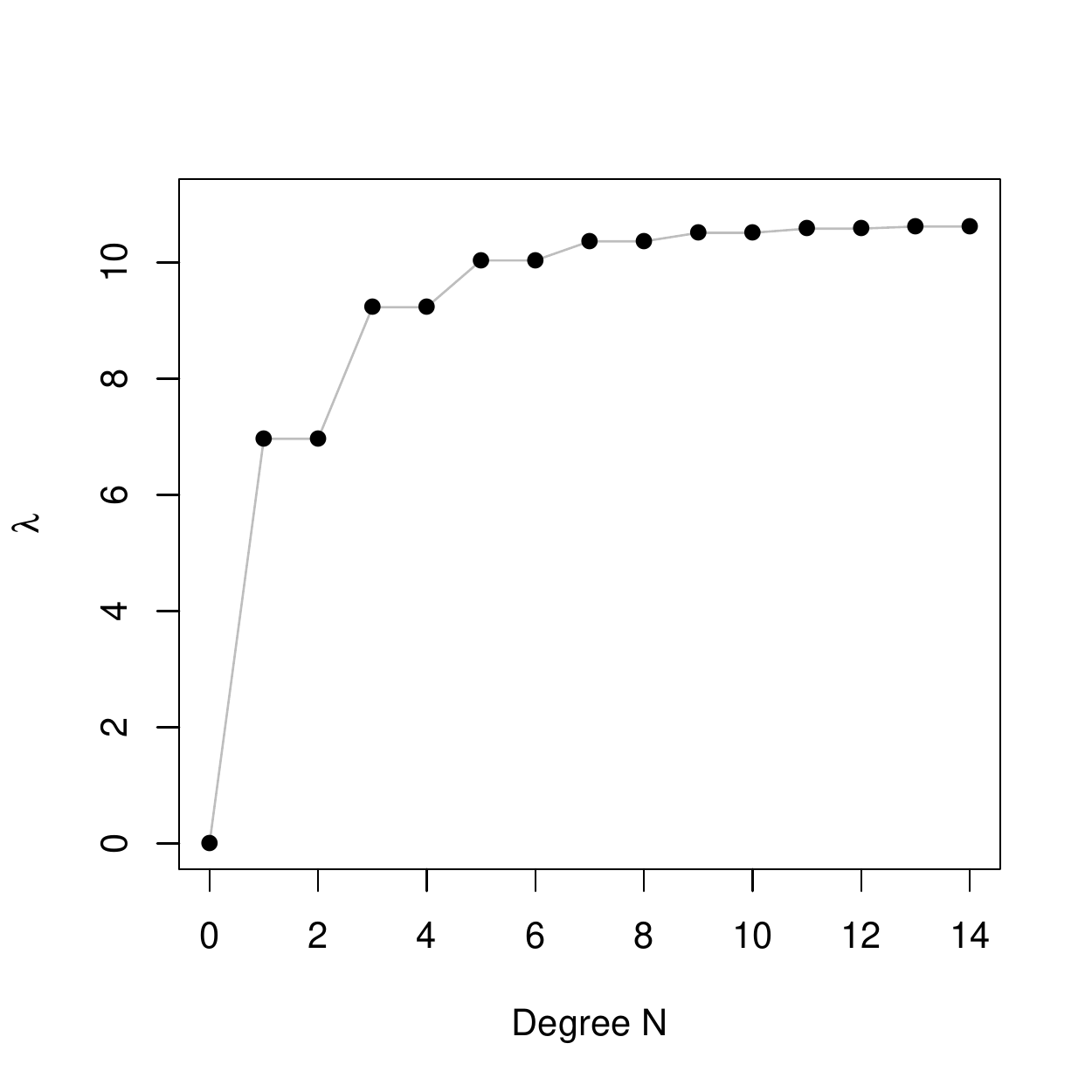}
  \caption{Scenario 1}
  \label{}
\end{subfigure}%
\begin{subfigure}{.45\linewidth}
\centering
 \includegraphics[width=1\linewidth]{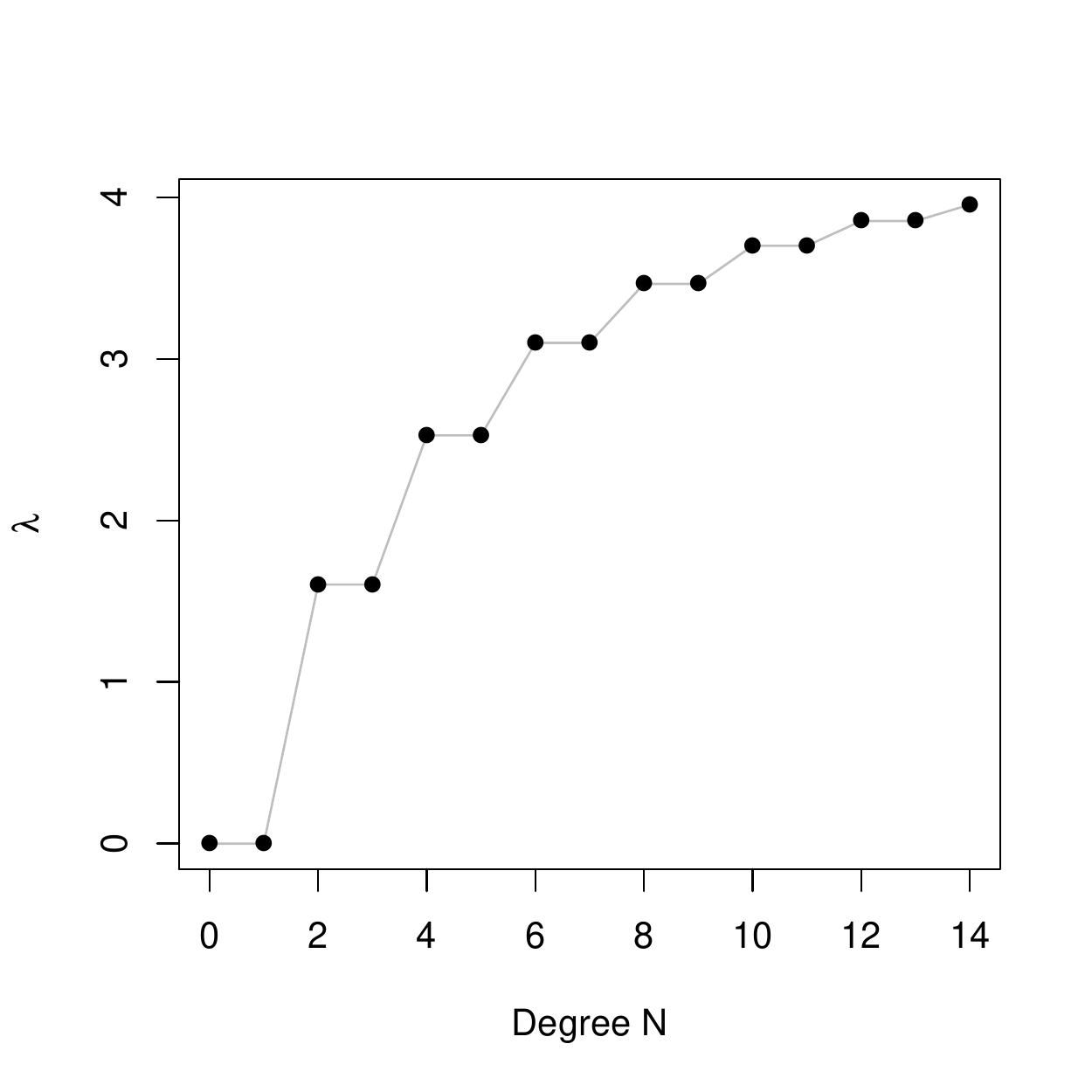}
  \caption{Scenario 2}
  \label{}
\end{subfigure}%
\caption{The ratio of between-class variation to within-class variation ($\lambda_N$) as a function of the truncation degree $N$ under (a) Scenario 1 and (b) Scenario 2.}\label{eigenincDet}
\end{figure}

\begin{figure}[ht!]
\centering
\begin{subfigure}{.4\linewidth}
\includegraphics[width=1\linewidth]{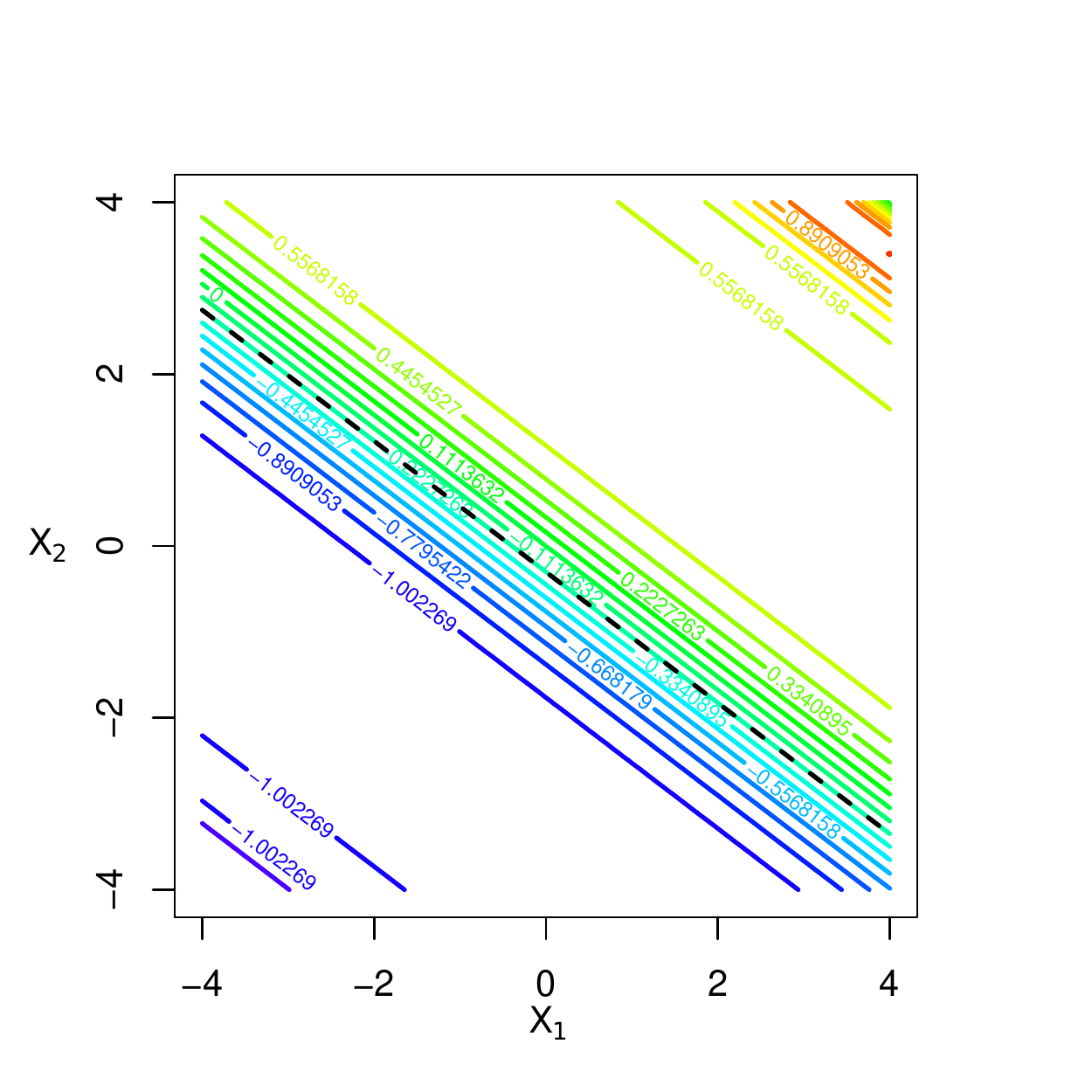}
 \caption{Scenario 1}
\end{subfigure}
\begin{subfigure}{.4\linewidth}
\includegraphics[width=1\linewidth]{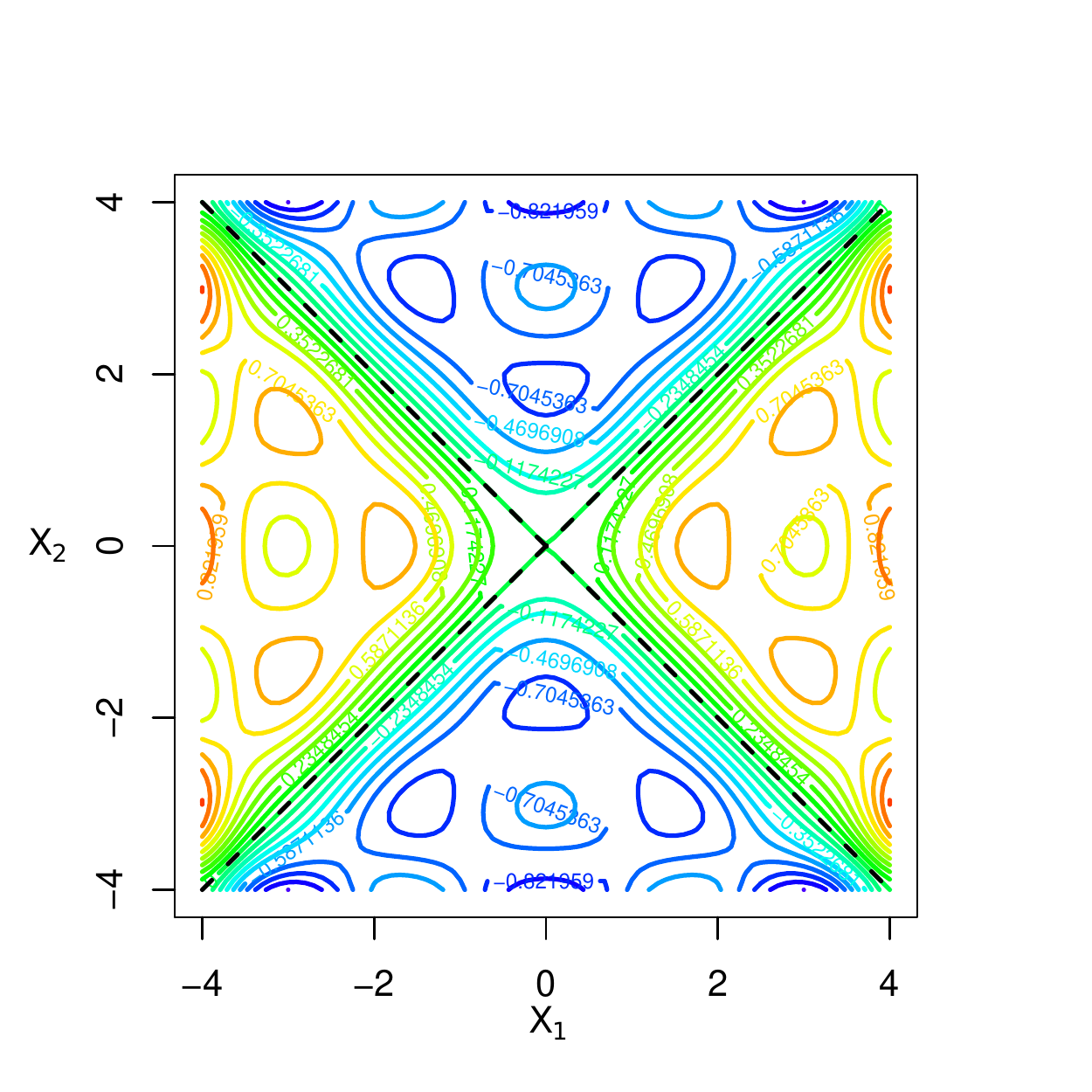}
  \caption{Scenario 2}
\end{subfigure}

\caption{Contours of the population Gaussian discriminants approximated by polynomials truncated at degree 14 under (a) Scenario 1 and  (b) Scenario 2. The black dashed lines are the optimal classification boundaries. }\label{contourforGauss}
\end{figure}

Figure~\ref{eigenincDet} shows how this eigenvalue $\lambda_N$ changes
with degree $N$ for each scenario. In Scenario 1, since a linear
component is essential, there is a sharp increase in $\lambda_N$ at
degree 1 followed by a gradual increase as odd features are added. By
contrast, in Scenario 2, $\lambda_N$ steadily increases as even
features are added. Overall the magnitude of the maximum ratio of
between-class variation to within-class variation ($\lambda_N$)
indicates that Scenario 1 presents an inherently easier problem than
Scenario 2. Figure~\ref{contourforGauss} displays some contours of the
approximate Gaussian discriminants for each scenario using $N=14$,
which suggest that the Gaussian kernel can capture the difference between classes effectively in both scenarios.

\noindent{\bf  Random Fourier feature representation:}
While polynomial features in the deterministic representation are naturally ordered by degree, there is no natural order in random Fourier features.
As with degree $N$ for deterministic features, however,  the Rayleigh
quotient as a measure of class separation or the corresponding eigenvalue increases as we add more random features. We numerically examine the effect of the number of random features $D$ on the eigenvalue $\lambda_D$ and monitor the increment in $\lambda_D$.

\begin{figure}[ht!]
  \vspace{-.3in}
  \centering
\begin{subfigure}{.45\linewidth}
\centering
 \includegraphics[width=1\linewidth]{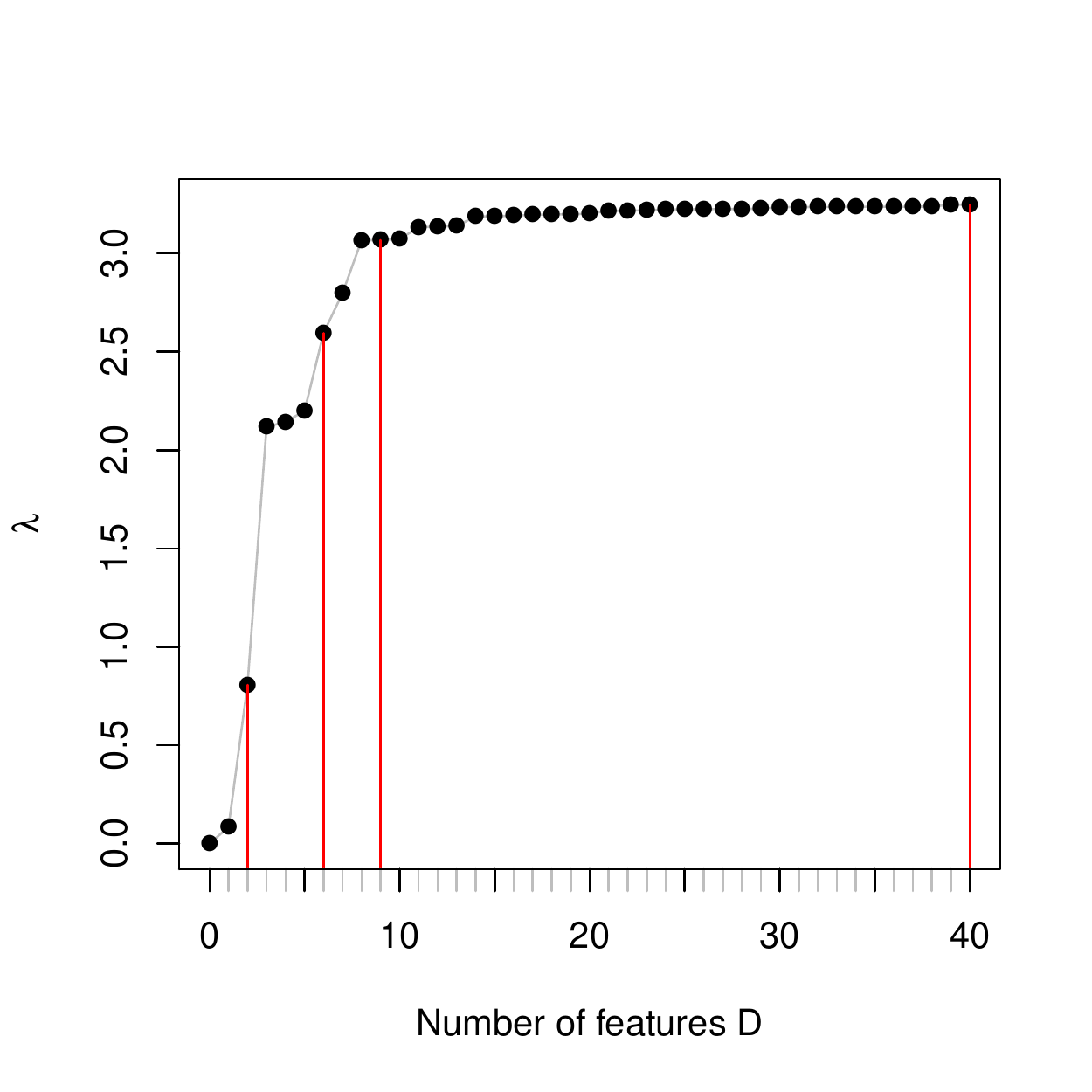}
  \caption{Scenario 1}
  \label{S1RFFEigen}
\end{subfigure}
\begin{subfigure}{.45\linewidth}
\centering
 \includegraphics[width=1\linewidth]{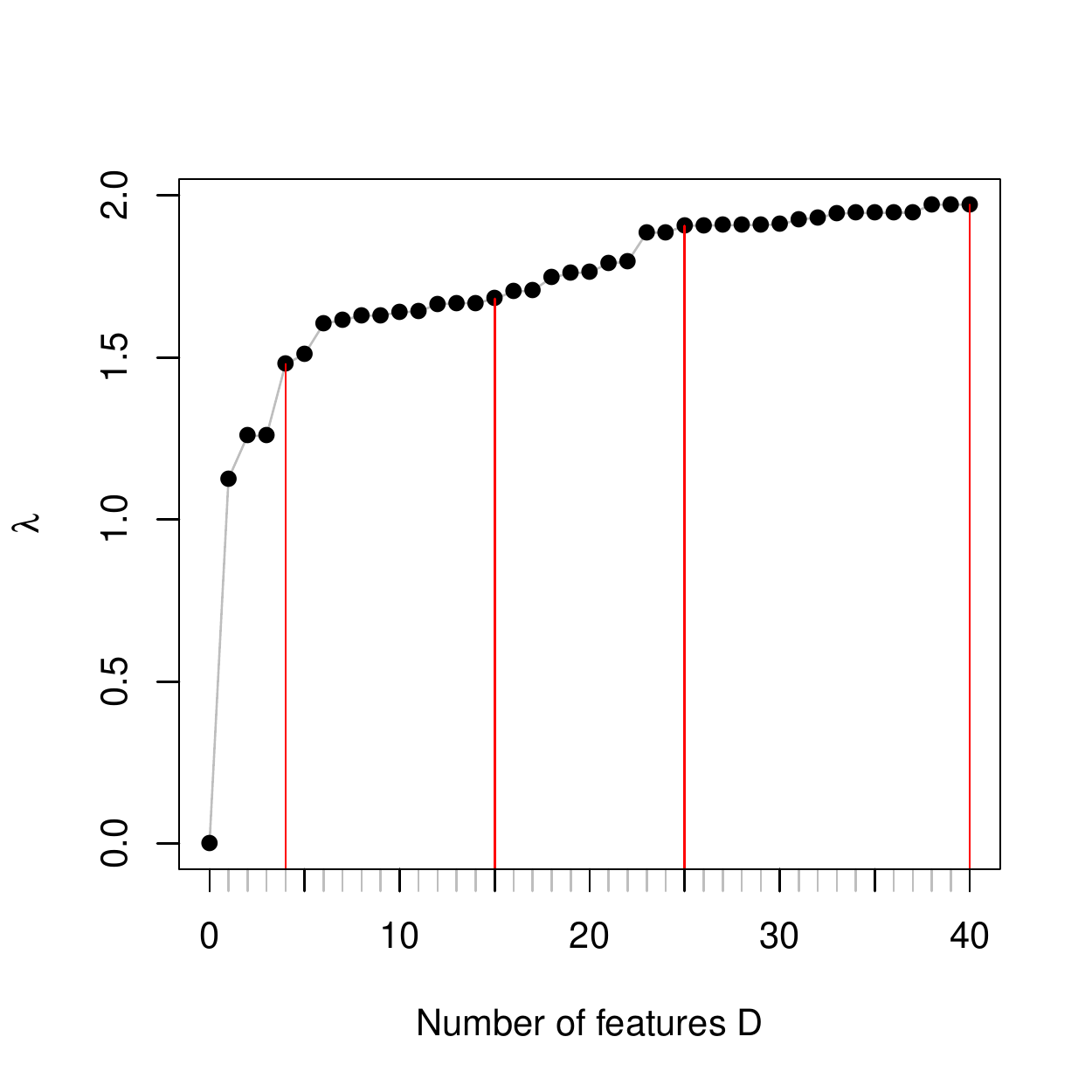}
  \caption{Scenario 2}
  \label{S2RFFEigen}
\end{subfigure}
\caption{The ratio of between-class variation to within-class variation ($\lambda_D$) as a function of the number of random Fourier features $D$ under (a) Scenario 1 and (b) Scenario 2. The red vertical lines indicate the number of random features used in Figures~\ref{RFFcontoursS1} and~\ref{RFFcontoursS2}. }\label{RFFeigen}
\end{figure}

\begin{figure}[ht!]
\includegraphics[width=1\linewidth]{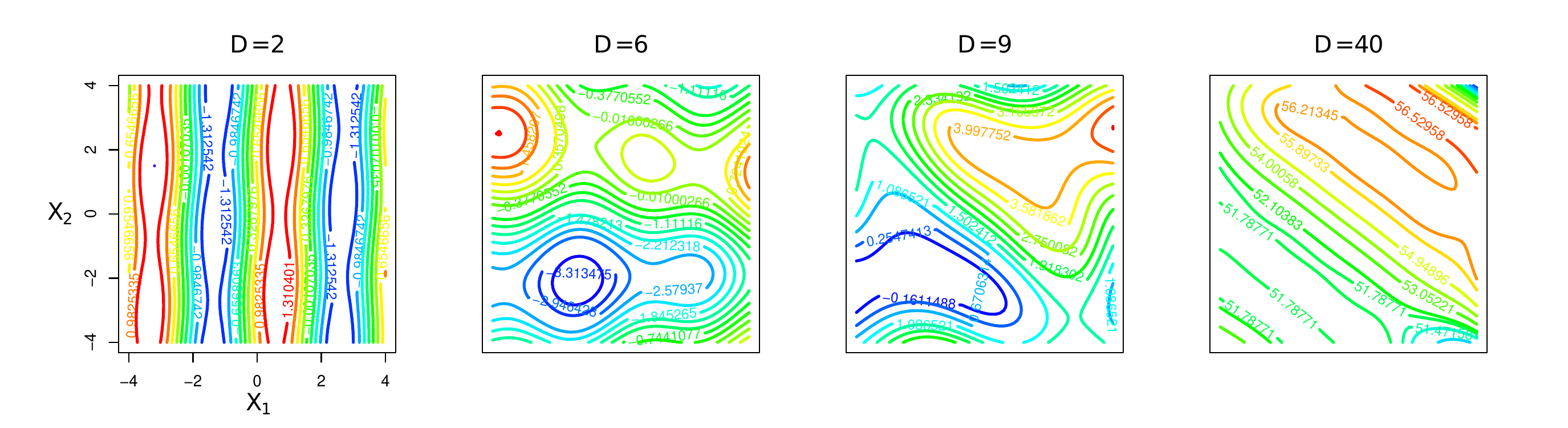}
\caption{Contours of the approximate discriminant functions using random Fourier features under Scenario 1.  The value of $D$ in each panel indicates the number of random Fourier features. }\label{RFFcontoursS1}
\end{figure}

\begin{figure}[ht!]
\includegraphics[width=1\linewidth]{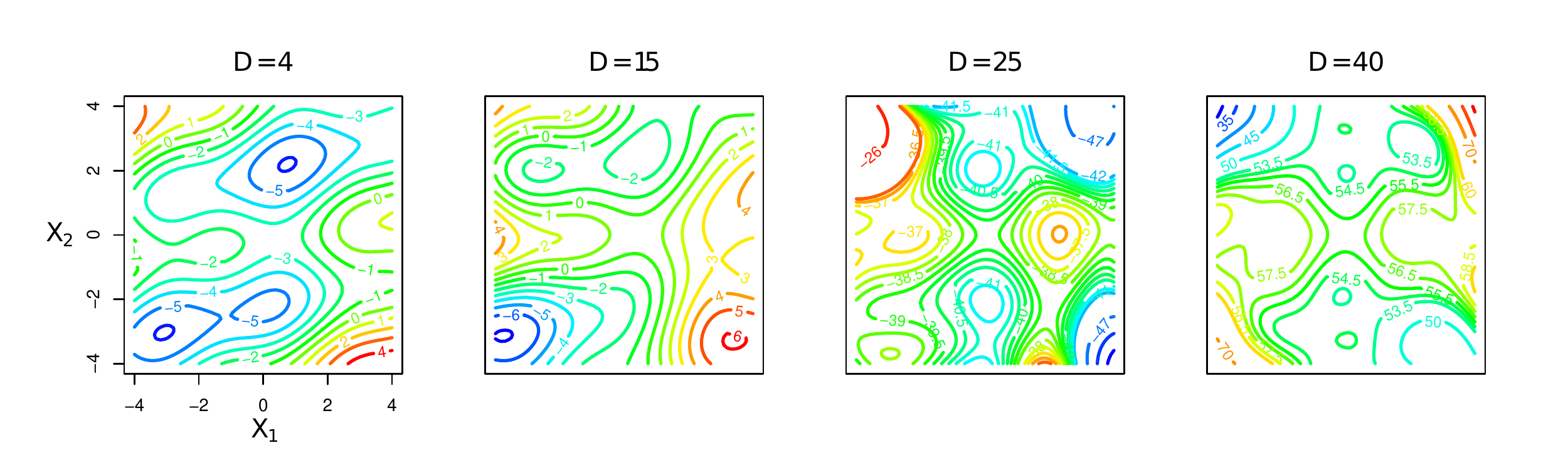}
\caption{Contours of the approximate discriminant functions using random Fourier features under Scenario 2.  The value of $D$ in each panel indicates the number of random Fourier features.}\label{RFFcontoursS2}
\end{figure}

For both scenarios, we randomly generated 40 $\w_i$ from $N_2(\0,
I_2)$ and $b_i$ from Uniform$(0, 2\pi)$, and defined phase-shifted cosine features, $z_{\w_i,b_i}(\x)=\sqrt{2}\cos(\w_i^\trans \x+b_i)$.
Figure~\ref{RFFeigen} shows how $\lambda_D$ changes with $D$ for each scenario.  Figure~\ref{RFFcontoursS1} shows how the approximate Gaussian discriminant in \eqref{RFFdisc} changes as the number of random features increases from 2 to 40 under Scenario 1. Figure~\ref{RFFcontoursS2} shows a similar change under Scenario 2. Those snapshots in Figures~\ref{RFFcontoursS1}  and~\ref{RFFcontoursS2} are chosen by monitoring the increment in the eigenvalue as more features are added. The number of features used is marked by the red vertical lines in Figure~\ref{RFFeigen} for reference.
As $D$ increases, the approximate Gaussian discriminants tend to better approximate the optimal classification boundaries.
Compared to the polynomial approximation, the eigenvalues level off
quickly with the number of random features $D$, and the maximum values
are far less than their counterparts with polynomial features in both
scenarios in part due to the randomness in the choice of  $\w_i$ and
$b_i$ and the fact that the nature of class difference is not harmonic. In summary,  
Fourier features are not as effective as polynomial features in these two settings.

\subsection{Real Data Example}\label{subsec:dataexample} 

In this section, we carry out a kernel discriminant analysis on the
spam email data set from the UCI Machine Learning Repository
\citep{Dua:2019}. We examine the geometry of sample kernel
discriminants with various kernels as in the simulation study, and
test the performance of the induced classifiers to see the impact of
the kernel choice and kernel parameters.

The data set contains information from 4601 email messages of which 
60.6\% are regular email and 39.4\% spam. The task is to detect whether a given email is regular or spam using 57 predictors available in order to filter out spam.
48 predictors are the percentage of words in the email that match a given word (e.g., credit, you, free), 6 predictors are the percentage of punctuation marks in the email that match a given punctuation mark (e.g., !, \$), and additional three predictors are
the longest, average, and total length of strings of capital letters in the message.

For ease of illustration,
we start with a low dimensional representation of the data using principal components
and construct kernel discriminants with those components rather than
the individual predictors. We observed that the predictors measuring
relative frequencies of words exhibit strong skewness in
distribution. To alleviate the skewness, we considered a logit
transformation before defining principal components. We also observed 
a large number of zeros on many predictors as some words do not
necessarily appear in every e-mail message. To handle this issue, we replaced zeros with a half of the least nonzero value in each predictor before taking a logit transformation and carried out a principal component analysis on the transformed data using their correlation matrix.
We then split the principal component scores into training and test sets of about 60\%  and 40\% each and evaluated the performance of trained classifiers over the test set.

\begin{figure}[ht!]
\centering
 \includegraphics[width=.6\linewidth]{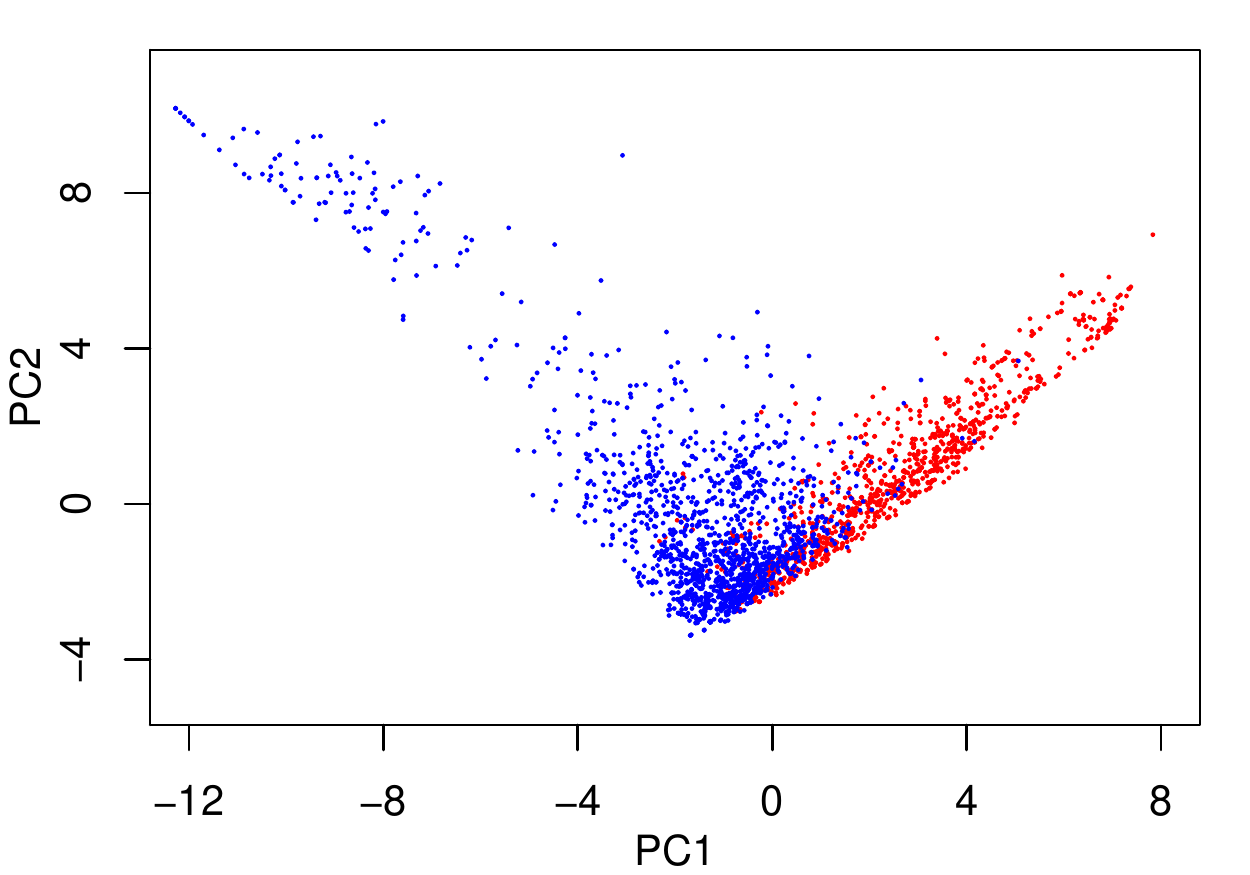}
  \caption{A scatterplot of the first two principal components scores on the email messages in the training data (blue: regular and red: spam). }
\label{fig:PCsforSpam}
\end{figure}

Figure~\ref{fig:PCsforSpam} 
shows the scores on the first two principal components for the training data. The two principal components explain 26\% of variation in the original data.
The score distributions for two types of email are skewed and substantially overlap with  very different covariances, suggesting that a nonlinear boundary is needed for classification.  

\begin{figure}[th!]
\centering
 \includegraphics[width=1\linewidth]{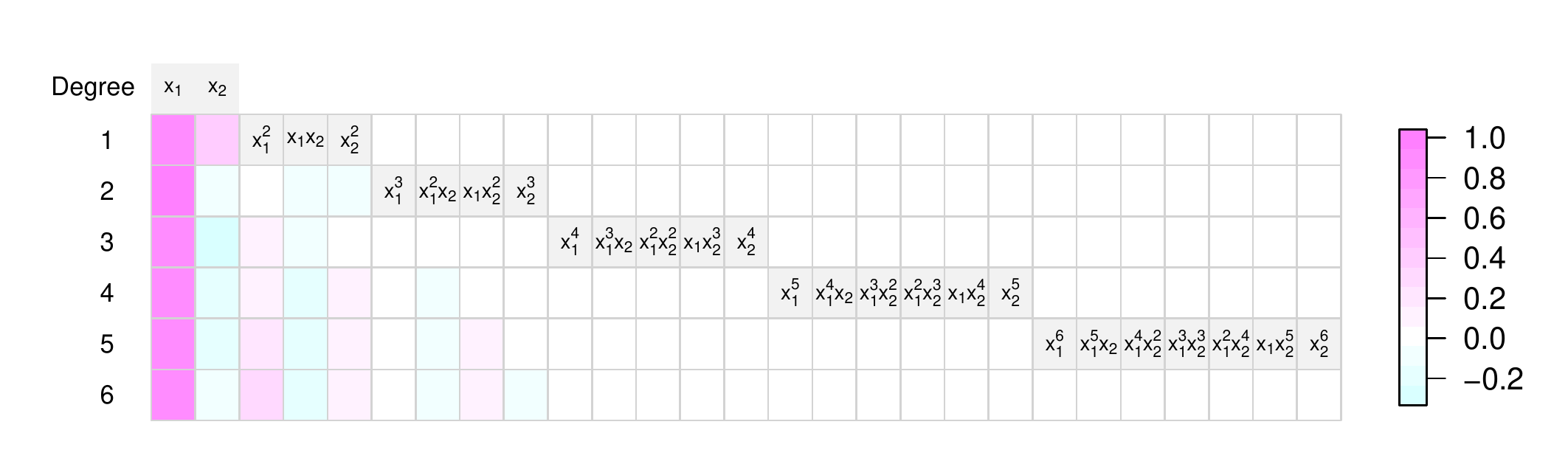}
  \caption{A color map of the estimated coefficients for the
    polynomial discriminants of degree 1 to 6 using two principal
    components from the spam email data displayed in the lower triangular array. The column label in the gray band (e.g., $x_1=PC_1$ and $x_2=PC_2$) indicates the term corresponding to each coefficient.}
\label{fig:coeffcolormap}
\end{figure}

\begin{figure}[ht!]
\centering
 \includegraphics[width=1\linewidth]{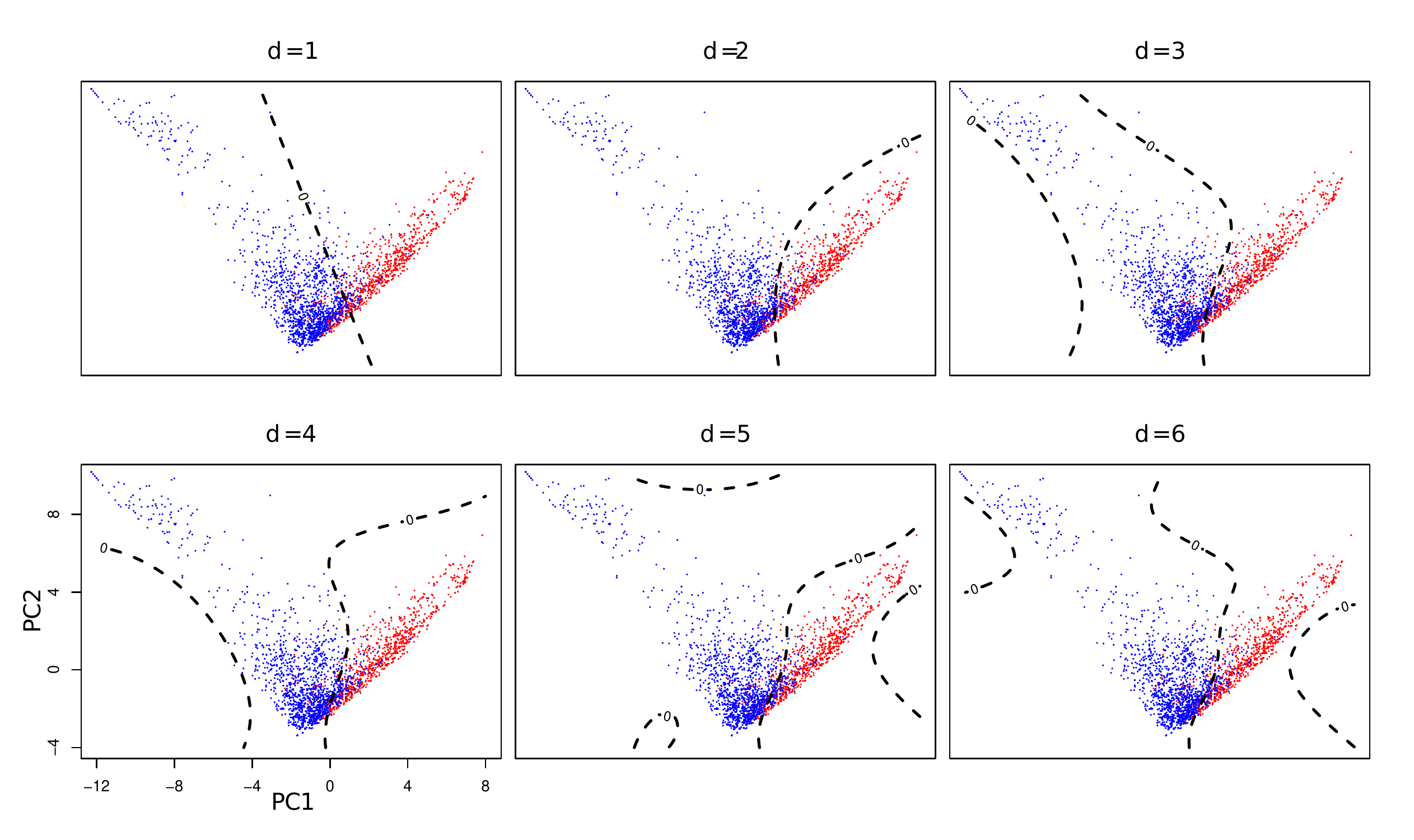}
  \caption{Decision boundaries of the polynomial discriminants with the inhomogeneous polynomial kernels of degree 1 to 6 obtained from the spam email data.  The black dashed lines are the boundaries with minimum training error for each kernel.}
\label{TheoreticalDiscSpamInHom}
\end{figure}

We performed a kernel discriminant analysis on the training data using the inhomogeneous polynomial kernels of degree 1 to 6, and obtained the corresponding polynomial discriminants. For computational efficiency, we estimated the moment difference $\tilde{\Delta}$ and covariance matrix $\tilde{\W}$ directly using the training data and solved a sample version of \eqref{eigeninhomo} instead of \eqref{BalphaWalpha}.
Figure~\ref{fig:coeffcolormap} shows the estimated coefficients for the discriminants that are normalized to unit length using a color map. High order terms, especially beyond the cubic terms, have negligible coefficients.
We need to decide on a threshold for discriminant scores to make a decision for spam filtering. We chose the threshold value by minimizing the training error. Figure~\ref{TheoreticalDiscSpamInHom} displays the decision boundaries of the final discriminant functions using the chosen threshold.  All nonlinear polynomial discriminants in the figure seem to have similar boundaries at least in the region where data density is high. 
Table~\ref{tab:spamtesterrorrates} presents their test error rates for comparison along with the rates for misclassifying spam as regular and vice versa.
The fifth and sixth order polynomial discriminants have the lowest error rate in this case. 
However, reduction in the test error rate is marginal after the third order, which we may expect from the result in Figure~\ref{fig:coeffcolormap} and diminishing returns in the ratio from degree as shown in Table~\ref{tab:spamtesterrorrates}.
We may well consider the cubic discriminant sufficient for this application.
It provides a good compromise between the two kinds of errors while maintaining simplicity.

\begin{table}[hb!]\centering
  \caption{Test error rates of kernel discriminant analysis on the spam email data set with the inhomogeneous polynomial kernels of varying degrees.
    The training error rates and  between-class to  within-class variation ratio are provided for comparison.  }\label{tab:spamtesterrorrates}
\begin{tabular}{c|c|c|ccc}
  \hline\hline
Degree               & Ratio & Training & \multicolumn{3}{c}{Test error} \\
                          &          & error      &Misclassified spam & Misclassified regular & Overall  \\[3pt]\hline
1                        & 4.3154 & 0.1381 &      0.2218      &      0.0744     & 0.1325         \\ 
2                        & 6.4226 & 0.1163 &     0.1887       &      0.0645     &  0.1135      \\
3                        & 7.2928 & 0.1116 &       0.1736     &       0.0645    & 0.1075         \\ 
4                        & 7.7043 & 0.1095 &       0.1377     &       0.0959    & 0.1124    \\ 
5                        & 8.0941 & 0.1058 &       0.1612     &       0.0672    & 0.1042    \\ 
6                        & 8.3192 & 0.1033 &       0.1543     &       0.0717    & 0.1042    \\ \hline
\end{tabular}
\end{table}

\section{Discussion} \label{sec:disc}

We have examined the population version of kernel discriminant analysis
and the generalized eigenvalue problem
with between-class and within-class kernel covariance operators 
to shed light on the relation
between the data distribution and resulting kernel discriminant.
Our analysis shows that polynomial discriminants  capture the difference between two
distributions through their moments of a certain order specified by the
polynomial kernel.
Depending on the representation of the Gaussian kernel, on the other
hand, Gaussian discriminants encode the class difference using
all polynomial features or Fourier features of random projections. 

Whenever we have some discriminative predictors in the data by
design as is typically the case, kernels of a simple form aligned
with those predictors will work well. For instance, if we use 
polynomial kernels in such a setting, we expect the Rayleigh quotient
as a measure of class separation to become saturated quickly with degree and 
low-order polynomial features to prevail.
The geometric perspective of kernel discriminant analysis presented in
this paper suggests that the ideal kernel for discrimination retains
only those features necessary for describing the difference in
two distributions.
This promotes a compositional view of kernels
(e.g., $\tilde{K}_d(\x,\u)=\sum_{m=0}^d{d\choose m}K_m(\x,\u)$) and further points to
the potential benefits of selecting kernel components 
relevant to discrimination similar to the way feature selection is
incorporated into linear discriminant analysis using sparsity inducing
penalties \citep{Clemmensen:etal2011, Cai:Liu2011}.
For instance, \cite{kim2006optimal} formulated a convex optimization problem for
kernel selection in KDA. It is also of interest to compare this kernel
selection approach with other approaches for numerical approximation
of kernel matrices themselves through Nystr\"{o}m approximation
\citep{Williams:Seeger2001, Drineas:Mahoney2005} or random projections \citep{ye2017fast}.

As a related issue, it has not been formally examined how the Rayleigh
quotient maximized in kernel discriminant analysis
is related to the error rate of the induced classifier except for some
special cases only.
It is of particular interest how the relation changes with the form of
a kernel and associated features given the difference between two distributions.

While our analysis has focused on the case of two classes, we can
generalize it to the case of multiple classes where more than one kernel
discriminants need to be considered and properly combined to make a decision.
We leave this extension as future research.

\section*{Acknowledgements}
This research was supported in part by the National Science Foundation
under grant DMS-15-13566.
We thank Professor Mikyoung Lim at KAIST for helpful conversations on linear operators.

\bibliography{Paperbib}
\bibliographystyle{dcu}

\end{document}